\documentclass[11pt,a4paper]{article}
\usepackage[hyperref]{aacl-ijcnlp2020}
\usepackage{times}
\usepackage{latexsym}

\usepackage{microtype}
\usepackage{graphicx}
\usepackage{url}
\usepackage{CJKutf8}
\usepackage{amsmath}
\usepackage{bm}

\usepackage{subfigure}
\usepackage{epsfig}
\usepackage{amssymb}
\usepackage{multirow}
\usepackage{xcolor}
\usepackage{color}
\usepackage{comment}
\usepackage{booktabs}

\aclfinalcopy 
\def\aclpaperid{36} 

\newcommand\SIZE{0.85}

\title{More Data, More Relations, More Context and More Openness:\\
A Review and Outlook for Relation Extraction }

\author{Xu Han$^{1}\thanks{\quad indicates equal contribution}$\hspace{0.5em}, Tianyu Gao$^{2*}$, Yankai Lin$^{3*}$, Hao Peng$^{1}$, Yaoliang Yang$^{1}$, Chaojun Xiao$^{1}$, \\ \textbf{Zhiyuan Liu}$^{1}$\thanks{\quad Corresponding author e-mail: liuzy@tsinghua.edu.cn}\hspace{0.5em}, \textbf{Peng Li}$^{3}$,  \textbf{Maosong Sun$^{1}$, Jie Zhou$^{3}$}\\
$^1$Department of Computer Science and Technology, Tsinghua University, Beijing, China\\
$^2$Princeton University, Princeton, NJ, USA\\
$^3$Pattern Recognition Center, WeChat AI, Tencent Inc, China\\
\tt hanxu17@mails.tsinghua.edu.cn, tianyug@princeton.edu, \\ 
\tt yankailin@tencent.com
}

\date{}

\begin{document}
\maketitle
\begin{abstract}
Relational facts are an important component of human knowledge, which are hidden in vast amounts of text. In order to extract these facts from text, people have been working on relation extraction (RE) for years. From early pattern matching to current neural networks, existing RE methods have achieved significant progress. Yet with explosion of Web text and emergence of new relations, human knowledge is increasing drastically, and we thus require ``more'' from RE: a more powerful RE system that can robustly utilize more data, efficiently learn more relations, easily handle more complicated context, and flexibly generalize to more open domains. In this paper, we look back at existing RE methods, analyze key challenges we are facing nowadays, and show promising directions towards more powerful RE. We hope our view can advance this field and inspire more efforts in the community.\footnote{Most of the papers mentioned in this work are collected into the following paper list~\url{https://github.com/thunlp/NREPapers}.}
\end{abstract}

\section{Introduction}

Relational facts organize knowledge of the world in a triplet format. These structured facts act as an import role of human knowledge and are explicitly or implicitly hidden in the text. For example, ``Steve Jobs co-founded Apple'' indicates the fact (\emph{Apple Inc.}, \texttt{founded by}, \emph{Steve Jobs}), and we can also infer the fact (\emph{USA}, \texttt{contains}, \emph{New York}) from ``Hamilton made its debut in New York, USA''. 

As these structured facts could benefit downstream applications, e.g, knowledge graph completion~\cite{bordes2013translating,wang2014knowledge}, search engine~\cite{xiong2017explicit,schlichtkrull2018modeling} and question answering~\cite{bordes2014question,dong2015question}, many efforts have been devoted to researching \textbf{relation extraction (RE)}, which aims at extracting relational facts from plain text. More specifically, after identifying entity mentions (e.g., \emph{USA} and \emph{New York}) in text, the main goal of RE is to classify relations (e.g., \texttt{contains}) between these entity mentions from their context. 

The pioneering explorations of RE lie in statistical approaches, such as pattern mining \citep{huffman1995learning,califf-mooney-1997-relational}, feature-based methods \citep{kambhatla2004combining} and graphical models \citep{roth2002probabilistic}. Recently, with the development of deep learning, neural models have been widely adopted for RE \citep{zeng2014relation,zhang2015bidirectional} and achieved superior results. These RE methods have bridged the gap between unstructured text and structured knowledge, and shown their effectiveness on several public benchmarks.

Despite the success of existing RE methods, most of them still work in a simplified setting. These methods mainly focus on training models with \textbf{large amounts} of \textbf{human annotations} to classify two given entities \textbf{within one sentence} into \textbf{pre-defined relations}. However, the real world is much more complicated than this simple setting: (1) collecting high-quality human annotations is expensive and time-consuming, (2) many long-tail relations cannot provide large amounts of training examples, (3) most facts are expressed by long context consisting of multiple sentences, and moreover (4) using a pre-defined set to cover those relations with open-ended growth is difficult. Hence, to build an effective and robust RE system for real-world deployment, there are still some more complex scenarios to be further investigated.



In this paper, we review existing RE methods (Section~\ref{sec:background}) as well as latest RE explorations (Section~\ref{sec:direction}) targeting more complex RE scenarios. Those feasible approaches leading to better RE abilities still require further efforts, and here we summarize them into four directions:

(1) \textbf{Utilizing More Data} (Section~\ref{sec:moredata}). Supervised RE methods heavily rely on expensive human annotations, while distant supervision \citep{mintz2009distant} introduces more auto-labeled data to alleviate this issue. Yet distant methods bring noise examples and just utilize single sentences mentioning entity pairs, which significantly weaken extraction performance. Designing schemas to obtain high-quality and high-coverage data to train robust RE models still remains a problem to be explored.

(2) \textbf{Performing More Efficient Learning} (Section~\ref{sec:morerelation}). Lots of long-tail relations only contain a handful of training examples. However, it is hard for conventional RE methods to well generalize relation patterns from limited examples like humans. Therefore, developing efficient learning schemas to make better use of limited or few-shot examples is a potential research direction.

(3) \textbf{Handling More Complicated Context} (Section~\ref{sec:morecontext}). Many relational facts are expressed in complicated context (e.g. multiple sentences or even documents), while most existing RE models focus on extracting intra-sentence relations. To cover those complex facts, it is valuable to investigate RE in more complicated context.

(4) \textbf{Orienting More Open Domains} (Section~\ref{sec:moreopen}). New relations emerge every day from different domains in the real world, and thus it is hard to cover all of them by hand. However, conventional RE frameworks are generally designed for pre-defined relations. Therefore, how to automatically detect undefined relations in open domains remains an open problem.

Besides the introduction of promising directions, we also point out two key challenges for existing methods: (1) \textbf{learning from text or names} (Section \ref{sec:name}) and (2) \textbf{datasets towards special interests} (Section \ref{sec:special}). We hope that all these contents could encourage the community to make further exploration and breakthrough towards better RE.

\section{Background and Existing Work}
\label{sec:background}


Information extraction (IE) aims at extracting structural information from unstructured text, which is an important field in natural language processing (NLP). Relation extraction (RE), as an important task in IE, particularly focuses on extracting relations between entities. A complete relation extraction system consists of a named entity recognizer to identify named entities (e.g., people, organizations, locations) from text, an entity linker to link entities to existing knowledge graphs (KGs, necessary when using relation extraction for knowledge graph completion), and a relational classifier to determine relations between entities by given context. 

\begin{figure}
    \centering
    \includegraphics[width = \SIZE\linewidth]{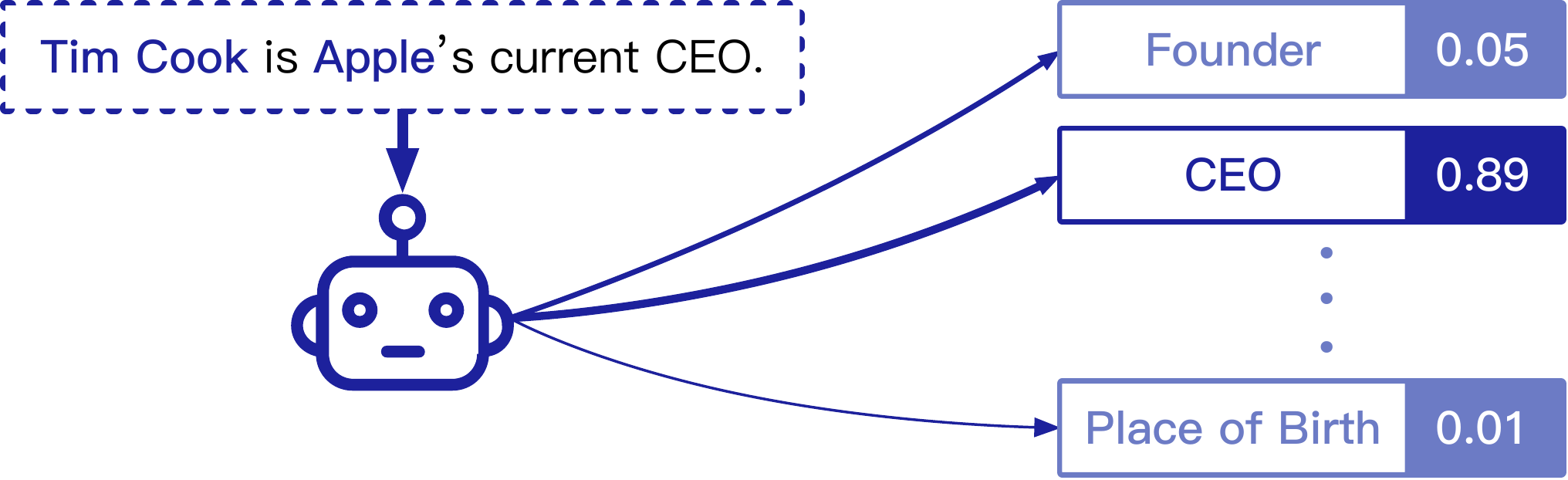}
    \caption{An example of RE. Given two entities and one sentence mentioning them, RE models classify the relation between them within a pre-defined relation set.}
    \label{fig:supervisedre}
\end{figure}

Among these steps, identifying the relation is the most crucial and difficult task, since it requires models to well understand the semantics of the context. Hence, RE generally focuses on researching the classification part, which is also known as relation classification. As shown in Figure~\ref{fig:supervisedre}, a typical RE setting is that given a sentence with two marked entities, models need to classify the sentence into one of the pre-defined relations\footnote{Sometimes there is a special class in the relation set indicating that the sentence does not express any pre-specified relation (usually named as N/A).}. 

In this section, we introduce the development of RE methods following the typical supervised setting, from early pattern-based methods, statistical approaches, to recent neural models.


\subsection{Pattern Extraction Models}

The pioneering methods use sentence analysis tools to identify syntactic elements in text, then automatically construct pattern rules from these elements~\cite{soderland1995crystal,kim1995acquisition,huffman1995learning,califf-mooney-1997-relational}. In order to extract patterns with better coverage and accuracy, later work involves larger corpora \citep{carlson2010toward}, more formats of patterns \citep{nakashole-etal-2012-patty,jiang2017metapad}, and more efficient ways of extraction~\citep{zheng-etal-2019-diag}. As automatically constructed patterns may have mistakes, most of the above methods require further examinations from human experts, which is the main limitation of pattern-based models.


\subsection{Statistical Relation Extraction Models}

As compared to using pattern rules, statistical methods bring better coverage and require less human efforts. Thus statistical relation extraction (SRE) has been extensively studied.

One typical SRE approach is \textbf{feature-based methods}~\cite{kambhatla2004combining,guodong2005exploring,jiang2007systematic,nguyen2007relation}, which design lexical, syntactic and semantic features for entity pairs and their corresponding context, and then input these features into relation classifiers.

Due to the wide use of support vector machines (SVM), \textbf{kernel-based methods} have been widely explored, which design kernel functions for SVM to measure the similarities between relation representations and textual instances~\cite{culotta2004dependency,bunescu2005shortest,zhao2005extracting,mooney2006subsequence,zhang2006composite,zhang2006exploring,wang2008re}.


There are also some other statistical methods focusing on extracting and inferring the latent information hidden in the text. \textbf{Graphical methods}~\cite{roth2002probabilistic,roth2004linear,sarawagi2005semi,yu2010jointly} abstract the dependencies between entities, text and relations in the form of directed acyclic graphs, and then use inference models to identify the correct relations. 

Inspired by the success of \textbf{embedding models} in other NLP tasks~\cite{mikolov2013efficient,mikolov2013distributed}, there are also efforts in encoding text into low-dimensional semantic spaces and extracting relations from textual embeddings~\cite{weston2013connecting,riedel2013relation,gormley2015improved}. Furthermore, \newcite{bordes2013translating},\newcite{wang2014knowledge} and \newcite{lin2015learning} utilize KG embeddings for RE.

Although SRE has been widely studied, it still faces some challenges. Feature-based and kernel-based models require many efforts to design features or kernel functions. While graphical and embedding methods can predict relations without too much human intervention, they are still limited in model capacities. There are some surveys systematically introducing SRE models~\cite{zelenko2003kernel,bach2007review,pawar2017relation}. In this paper, we do not spend too much space for SRE and focus more on neural-based models.

\subsection{Neural Relation Extraction Models}

Neural relation extraction (NRE) models introduce neural networks to automatically extract semantic features from text. Compared with SRE models, NRE methods can effectively capture textual information and generalize to wider range of data.


Studies in NRE mainly focus on designing and utilizing various network architectures to capture the relational semantics within text, such as \textbf{recursive neural networks}~\citep{socher2012semantic,miwa2016end} that learn compositional representations for sentences recursively, 
\textbf{convolutional neural networks (CNNs)}~\citep{liu2013convolution,zeng2014relation,santos2015classifying,nguyen2015relation,zeng2015distant,huang2017deep} that effectively model local textual patterns, \textbf{recurrent neural networks (RNNs)}~\citep{zhang2015relation,nguyen2015combining,vu2016combining,zhang2015bidirectional} that can better handle long sequential data, \textbf{graph neural networks (GNNs)}~\citep{zhang2018graph,zhu2019graph} that build word/entity graphs for reasoning, and \textbf{attention-based neural networks}~\citep{zhou2016attention,wang2016relation,xiao2016semantic} that utilize attention mechanism to aggregate global relational information. 

\begin{figure}[t]
    \centering
    \includegraphics[width = \SIZE\linewidth]{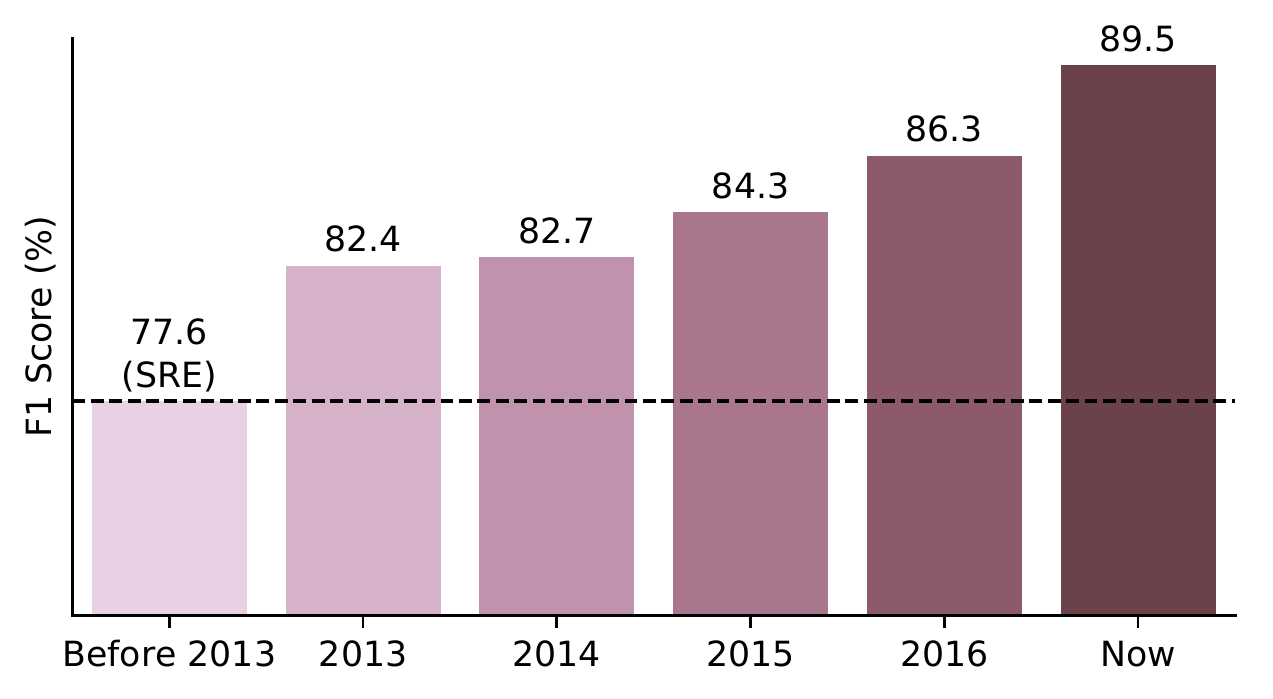}
    \caption{The performance of state-of-the-art RE models in different years on widely-used dataset SemEval-2010 Task 8. The adoption of neural models (since 2013) has brought great improvement in performance.}
    \label{fig:semeval}
\end{figure}

Different from SRE models, NRE mainly utilizes word embeddings and position embeddings instead of hand-craft features as inputs. \textbf{Word embeddings}~\cite{turian2010word,mikolov2013distributed} are the most used input representations in NLP, which encode the semantic meaning of words into vectors. In order to capture the entity information in text, \textbf{position embeddings}~\citep{zeng2014relation} are introduced to specify the relative distances between words and entities. Except for word embeddings and position embeddings, there are also other works integrating syntactic information into NRE models. \citet{xu2015semantic} and \citet{xu2015classifying} adopt CNNs and RNNs over \textbf{shortest dependency paths} respectively. \citet{liu2015dependency} propose a recursive neural network based on augmented dependency paths. \citet{xu2016improved} and \citet{cai2016bidirectional} utilize deep RNNs to make further use of dependency paths. Besides, there are some efforts combining NRE with \textbf{universal schemas}~\cite{verga2016multilingual,vergamccallum,riedel2013relation}. Recently, \textbf{Transformers} \citep{vaswani2017attention} and \textbf{pre-trained language models}~\cite{devlin2019bert} have also been explored for NRE~\cite{du2018multi,verga2018simultaneously,wu2019enriching,soares2019matching} and have achieved new state-of-the-arts.

By concisely reviewing the above techniques, we are able to track the development of RE from pattern and statistical methods to neural models. Comparing the performance of state-of-the-art RE models in years (Figure~\ref{fig:semeval}), we can see the vast increase since the emergence of NRE, which demonstrates the power of neural methods.

\section{``More'' Directions for RE}
\label{sec:direction}

Although the above-mentioned NRE models have achieved superior results on benchmarks, they are still far from solving the problem of RE. Most of these models utilize abundant human annotations and just aim at extracting pre-defined relations within single sentences. Hence, it is hard for them to work well in complex cases. In fact, there have been various works exploring feasible approaches that lead to better RE abilities on real-world scenarios. In this section, we summarize these exploratory efforts into four directions, and give our review and outlook about these directions.

\begin{figure}
    \centering
    \includegraphics[width = \SIZE\linewidth]{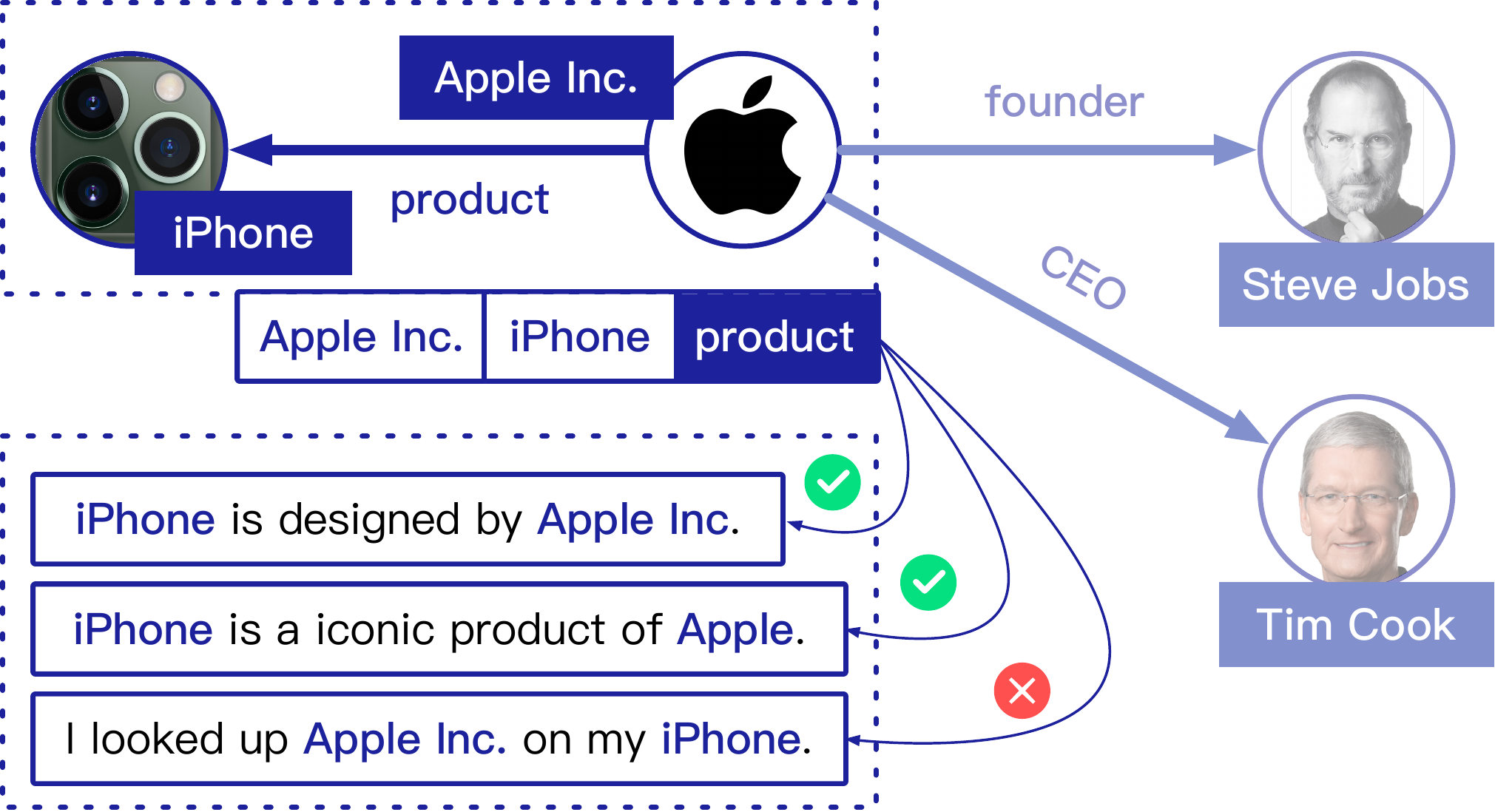}
    \caption{An example of distantly supervised relation extraction. With the fact (\emph{Apple Inc.}, \texttt{product}, \emph{iPhone}), DS finds all sentences mentioning the two entities and annotates them with the relation \texttt{product}, which inevitably brings noise labels.}
    \label{fig:distant}
\end{figure}

\subsection{Utilizing More Data}
\label{sec:moredata}

Supervised NRE models suffer from the lack of large-scale high-quality training data, since manually labeling data is time-consuming and human-intensive. To alleviate this issue, distant supervision (DS) assumption has been used to automatically label data by aligning existing KGs with plain text~\cite{mintz2009distant,nguyen2011end,min2013distant}. As shown in Figure~\ref{fig:distant}, for any entity pair in KGs, sentences mentioning both the entities will be labeled with their corresponding relations in KGs. Large-scale training examples can be easily constructed by this heuristic scheme. 

Although DS provides a feasible approach to utilize more data, this automatic labeling mechanism is inevitably accompanied by the wrong labeling problem. The reason is that not all sentences mentioning the two entities express their relations in KGs exactly. For example, we may mistakenly label ``\emph{Bill Gates} retired from \emph{Microsoft}'' with the relation \texttt{founder}, if (\emph{Bill Gates}, \texttt{founder}, \emph{Microsoft}) is a relational fact in KGs. 

\begin{table}
    \centering
    \small
    \begin{tabular}{c|cccc}
    \toprule
        Dataset & \#Rel. & \#Fact & \#Inst. & N/A\\
    \midrule
        NYT-10 & 53 & 377,980 & 694,491 & 79.43\%\\
        Wiki-Distant & 454 & 605,877 & 1,108,288 & 47.61\%\\
    \bottomrule
    \end{tabular}
    \caption{Statistics for NYT-10 and Wiki-Distant. Four columns stand for numbers of relations, facts and instances, and proportions of N/A instances respectively.}
    \label{tab:dataset}
\end{table}

\begin{table}
    \centering
    \small
    \begin{tabular}{c|cc}
        \toprule
        Model &  NYT-10 & Wiki-Distant \\
        \midrule
        PCNN-ONE & 0.340 & 0.214 \\
        PCNN-ATT & 0.349 & 0.222 \\
        BERT & 0.458 & 0.361 \\
        \bottomrule
    \end{tabular}
      \caption[Caption for LOF]{Area under the curve (AUC) of PCNN-ONE \citep{zeng2015distant}, PCNN-ATT \citep{lin2016neural} and BERT \citep{devlin2019bert} on two datasets.}
    \label{tab:distantexp}
\end{table}

The existing methods to alleviate the noise problem can be divided into three major approaches:

 (1) Some methods adopt multi-instance learning by combining sentences with same entity pairs and then \textbf{selecting informative instances} from them. \citet{riedel2010modeling,hoffmann2011knowledge,surdeanu2012multi} utilize graphical model to infer the informative sentences, while \citet{zeng2015distant} use a simple heuristic selection strategy. Later on, \citet{lin2016neural,zhang2017position,han2018hierarchical,li2020self,zhu2019improving,hu2019improving} design attention mechanisms to highlight informative instances for RE. 

 (2) \textbf{Incorporating extra context information} to denoise DS data has also been explored, such as incorporating KGs as external information to guide instance selection~\cite{ji2017distant,han2018neural,zhang2019long,qu2019fine} and adopting multi-lingual corpora for the information consistency and complementarity~\citep{verga2016multilingual,lin2017neural,wang2018adversarial}.

 (3) Many methods tend to utilize \textbf{sophisticated mechanisms and training strategies} to enhance distantly supervised NRE models. \citet{vu2016combining,beltagy2019combining} combine different architectures and training strategies to construct hybrid frameworks. \citet{liu2017soft} incorporate a soft-label scheme by changing unconfident labels during training. Furthermore, reinforcement learning~\cite{feng2018reinforcement,zeng2018large} and adversarial training~\cite{wu2017adversarial,wang2018adversarial,han2018denoising} have also been adopted in DS. 

\begin{figure}[t]
    \centering
    \includegraphics[width=0.49\linewidth]{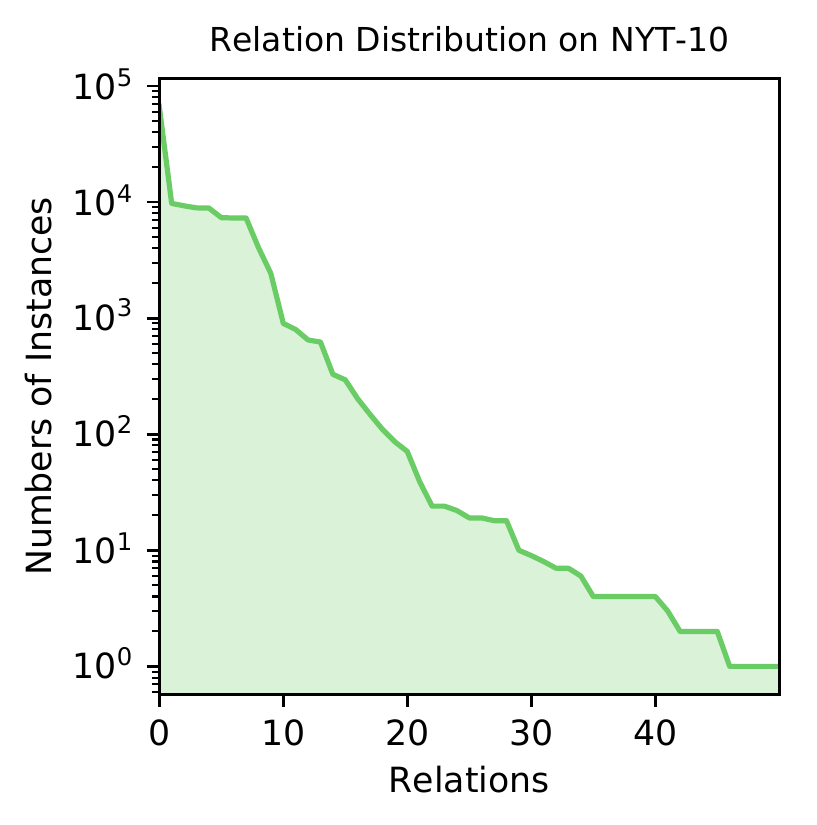}
    \includegraphics[width=0.49\linewidth]{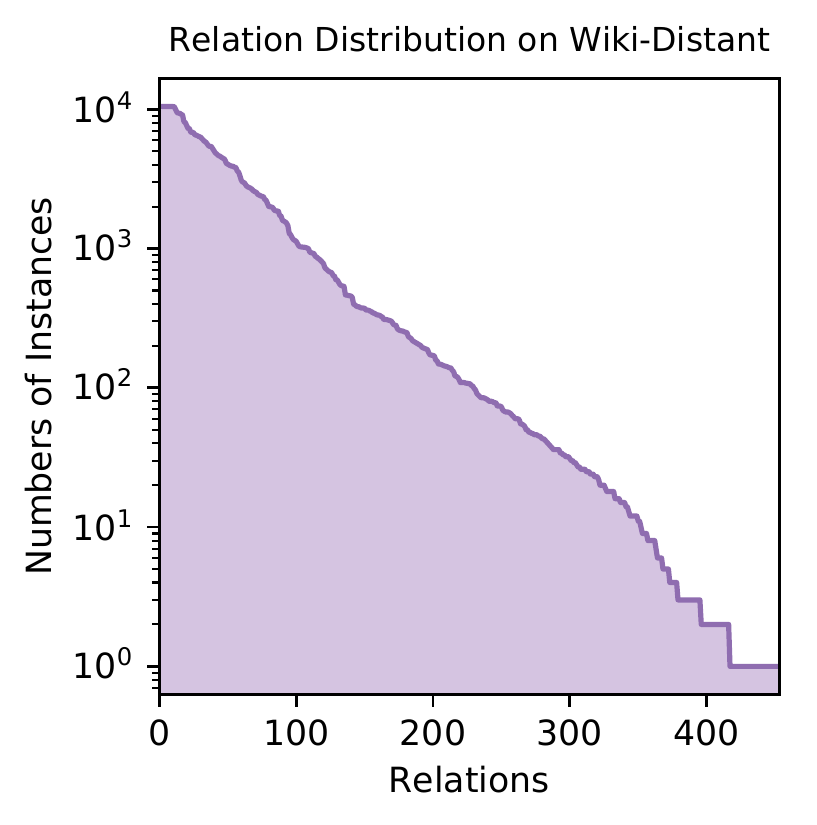}
    \caption{Relation distributions (log-scale) on the training part of DS datasets NYT-10 and Wiki-Distant, suggesting that real-world relation distributions suffer from the long-tail problem.}
    \label{fig:longtail}
\end{figure}

The researchers have formed a consensus that utilizing more data is a potential way towards more powerful RE models, and there still remains some open problems worth exploring:

(1) Existing DS methods focus on \textbf{denoising auto-labeled instances} and it is certainly meaningful to follow this research direction. Besides, current DS schemes are still similar to the original one in~\citep{mintz2009distant}, which just covers the case that the entity pairs are mentioned in the same sentences. To achieve better coverage and less noise, \textbf{exploring better DS schemes} for auto-labeling data is also valuable. 

(2) Inspired by recent work in adopting pre-trained language models~\cite{zhang2019ernie,wu2019enriching,soares2019matching} and active learning~\cite{zheng-etal-2019-diag} for RE, to \textbf{perform unsupervised or semi-supervised learning} for utilizing large-scale unlabeled data as well as using knowledge from KGs and introducing human experts in the loop is also promising.

Besides addressing existing approaches and future directions, we also propose a new DS dataset to advance this field, which will be released once the paper is published. The most used benchmark for DS, NYT-10 \citep{riedel2010modeling}, suffers from small amount of relations, limited relation domains and extreme long-tail relation performance. To alleviate these drawbacks, we utilize Wikipedia and Wikidata \citep{vrandevcic2014wikidata} to construct \textbf{Wiki-Distant} in the same way as \citet{riedel2010modeling}. As demonstrated in Table \ref{tab:dataset}, Wiki-Distant covers more relations and possesses more instances, with a more reasonable N/A proportion. Comparison results of state-of-the-art models on these two datasets\footnote{Due to the large size, we do not use any denoise mechanism for BERT, which still achieves the best results.} are shown in Table~\ref{tab:distantexp}, indicating that Wiki-Distant is more challenging and there is a long way to resolve distantly supervised RE.

\subsection{Performing More Efficient Learning}
\label{sec:morerelation}

\begin{figure}[t]
    \centering
    \includegraphics[width = \SIZE\linewidth]{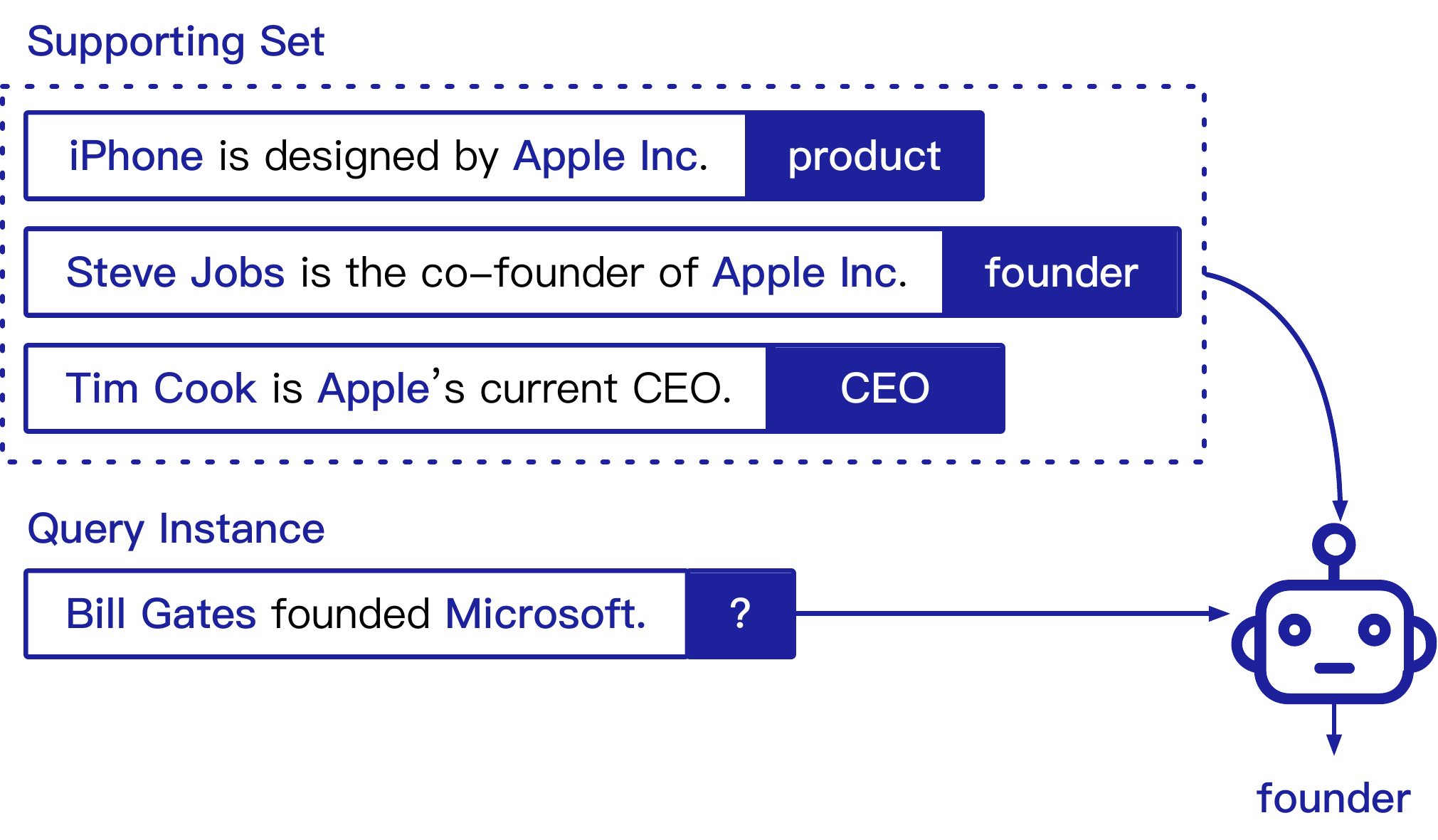}
    \caption{An example of few-shot RE. Give a few instances for new relation types, few-shot RE models classify query sentences into one of the given relations.}
    \label{fig:fewshot}
\end{figure}

Real-world relation distributions are long-tail: Only the common relations obtain sufficient training instances and most relations have very limited relational facts and corresponding sentences. We can see the long-tail relation distributions on two DS datasets from Figure \ref{fig:longtail}, where many relations even have less than 10 training instances.
This phenomenon calls for models that can learn long-tail relations more efficiently. Few-shot learning, which focuses on grasping tasks with only a few training examples, is a good fit for this need.

To advance this field, \citet{han2018fewrel} first built a large-scale few-shot relation extraction dataset (FewRel). This benchmark takes the $N$-way $K$-shot setting, where models are given $N$ random-sampled new relations, along with $K$ training examples for each relation. With limited information, RE models are required to classify query instances into given relations (Figure \ref{fig:fewshot}). 

The general idea of few-shot models is to train good representations of instances or learn ways of fast adaptation from existing large-scale data, and then transfer to new tasks. There are mainly two ways for handling few-shot learning: (1) \textbf{Metric learning} learns a semantic metric on existing data and classifies queries by comparing them with training examples \citep{koch2015siamese,vinyals2016matching,snell2017prototypical,soares2019matching}. While most metric learning models perform distance measurement on sentence-level representation, \citet{ye-ling-2019-multi,gao-etal-2019-fewrel} utilize token-level attention for finer-grained comparison. (2) \textbf{Meta-learning}, also known as ``learning to learn'', aims at grasping the way of parameter initialization and optimization through the experience gained on the meta-train data \citep{ravi2016optimization, finn2017model, mishra2018simple}.

Researchers have made great progress in few-shot RE.
However, there remain many challenges that are important for its applications and have not yet been discussed. \citet{gao-etal-2019-fewrel} propose two problems worth further investigation:

\begin{figure}
    \centering
    \includegraphics[width=0.96\linewidth]{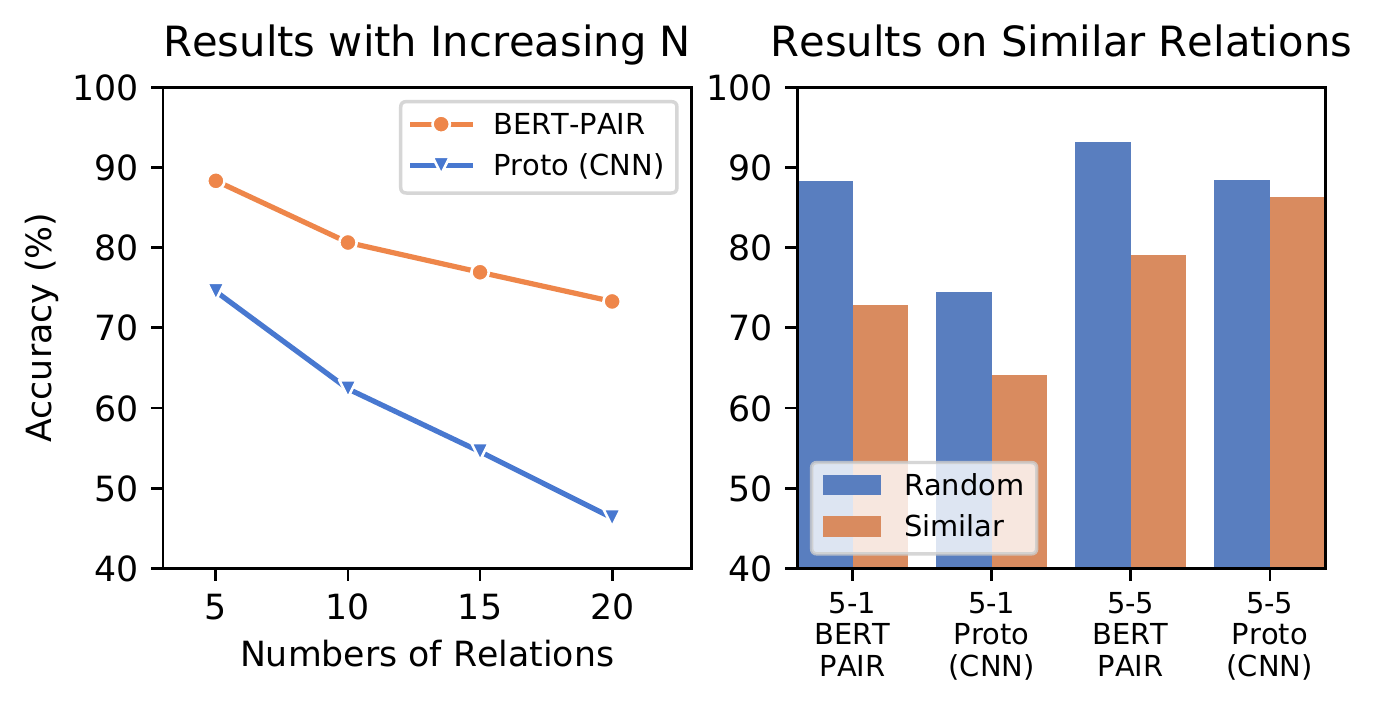}
    \caption{Few-shot RE results with (A) increasing $N$ and (B) similar relations. The left figure shows the accuracy (\%) of two models in $N$-way 1-shot RE. In the right figure, ``random'' stands for the standard few-shot setting and ``similar'' stands for evaluating with selected similar relations.}
    \label{fig:fewshotexp}
\end{figure}

(1) \textbf{Few-shot domain adaptation} studies how few-shot models can transfer across domains . It is argued that in the real-world application, the test domains are typically lacking annotations and could differ vastly from the training domains. Thus, it is crucial to evaluate the transferabilities of few-shot models across domains.

(2) \textbf{Few-shot none-of-the-above detection} is about detecting query instances that do not belong to any of the sampled $N$ relations. In the $N$-way $K$-shot setting, it is assumed that all queries express one of the given relations. However, the real case is that most sentences are not related to the relations of our interest. Conventional few-shot models cannot well handle this problem due to the difficulty to form a good representation for the none-of-the-above (NOTA) relation. Therefore, it is crucial to study how to identify NOTA instances.

(3) Besides the above challenges, it is also important to see that, the existing \textbf{evaluation protocol} may over-estimate the progress we made on few-shot RE. Unlike conventional RE tasks, few-shot RE randomly samples $N$ relations for each evaluation episode; in this setting, the number of relations is usually very small (5 or 10) and it is very likely to sample $N$ distinct relations and thus reduce to a very easy classification task.

We carry out two simple experiments to show the problems (Figure \ref{fig:fewshotexp}): (A) We evaluate few-shot models with increasing $N$ and the performance drops drastically with larger relation numbers. Considering that the real-world case contains much more relations, it shows that existing models are still far from being applied. (B) Instead of randomly sampling $N$ relations, we hand-pick $5$ relations similar in semantics and evaluate few-shot RE models on them. It is no surprise to observe a sharp decrease in the results, which suggests that existing few-shot models may overfit simple textual cues between relations instead of really understanding the semantics of the context. More details about the experiments are in Appendix A.

\subsection{Handling More Complicated Context}
\label{sec:morecontext}

\begin{figure}
    \centering
    \includegraphics[width = \SIZE\linewidth]{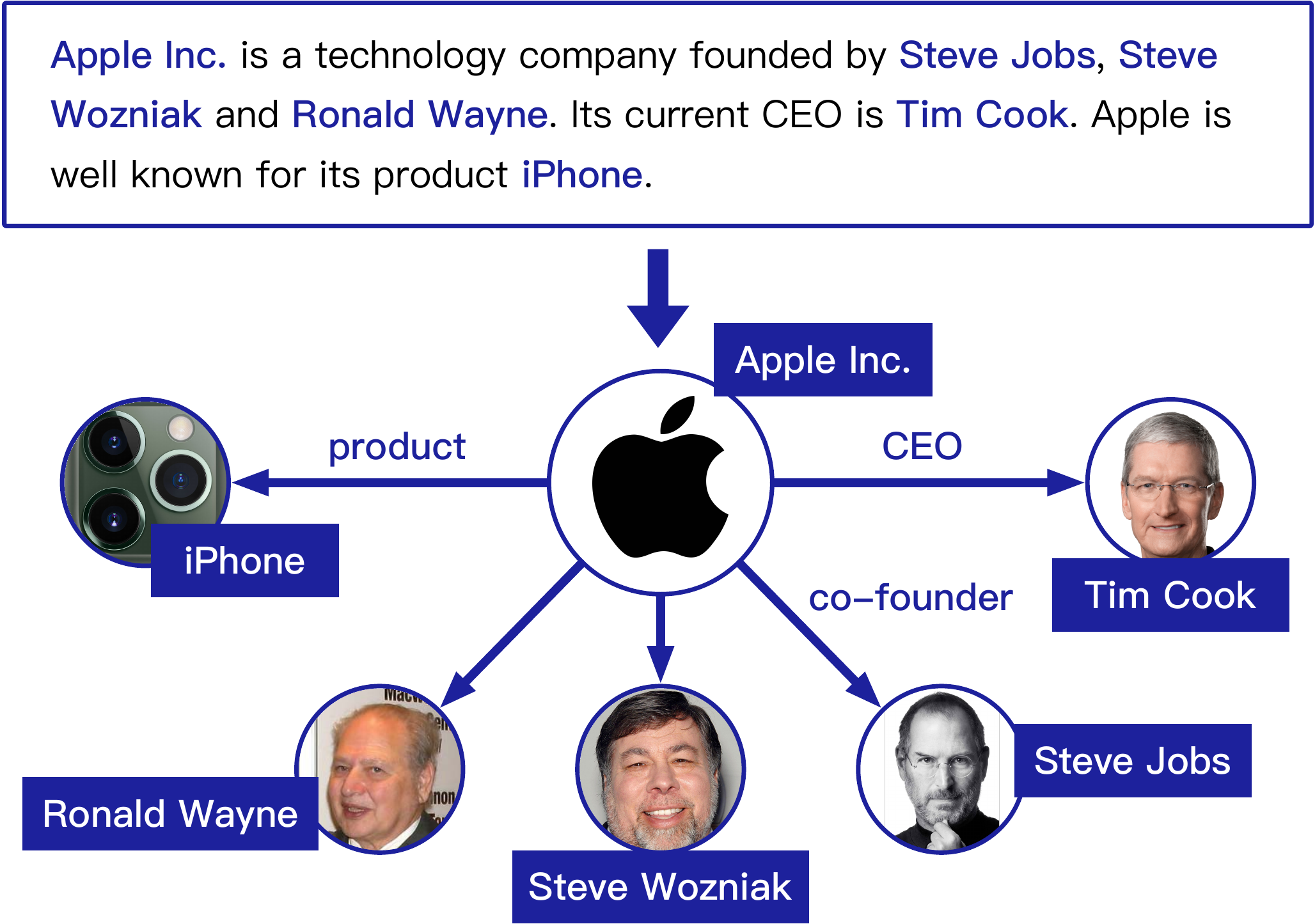}
    \caption{An example of document-level RE. Given a paragraph with several sentences and multiple entities, models are required to extract all possible relations between these entities expressed in the document.}
    \label{fig:document}
\end{figure}

As shown in Figure~\ref{fig:document}, one document generally mentions many entities exhibiting complex cross-sentence relations. Most existing methods focus on intra-sentence RE and thus are inadequate for collectively identifying these relational facts expressed in a long paragraph. In fact, most relational facts can only be extracted from complicated context like documents rather than single sentences~\cite{yao2019docred}, which should not be neglected.

There are already some works proposed to extract relations across multiple sentences: 

(1) \textbf{Syntactic methods}~\cite{wick2006EMNLP,gerber2010ACL,swampillai2011RANLP,yoshikawa2011,quirk2017distant} rely on textual features extracted from various syntactic structures, such as coreference annotations, dependency parsing trees and discourse relations, to connect sentences in documents.

(2) \citet{zeng2016incorporating,christopoulou2018walk} build inter-sentence entity graphs, which can utilize \textbf{multi-hop paths between entities} for inferring the correct relations.

(3) \newcite{peng2017nary,song2018EMNLP,zhu2019towards} employ \textbf{graph-structured neural networks} to model cross-sentence dependencies for relation extraction, which bring in memory and reasoning abilities.

To advance this field, some document-level RE datasets have been proposed. \citet{quirk2017distant,peng2017nary} build datasets by DS. \citet{li2016biocreative,peng2017nary} propose datasets for specific domains. \citet{yao2019docred} construct a general document-level RE dataset annotated by crowdsourcing workers, suitable for evaluating general-purpose document-level RE systems. 

Although there are some efforts investing into extracting relations from complicated context (e.g., documents), the current RE models for this challenge are still crude and straightforward. Followings are some directions worth further investigation:

(1) Extracting relations from complicated context is a challenging task requiring \textbf{reading, memorizing and reasoning} for discovering relational facts across multiple sentences. Most of current RE models are still very weak in these abilities. 

(2) Besides documents, \textbf{more forms of context} is also worth exploring, such as extracting relational facts across documents, or understanding relational information based on heterogeneous data.

(3) Inspired by~\newcite{narasimhan2016improving}, which utilizes search engines for acquiring external information,  \textbf{automatically searching and analysing context for RE} may help RE models identify relational facts with more coverage and become practical for daily scenarios.

\begin{figure}
    \centering
    \includegraphics[width = \SIZE\linewidth]{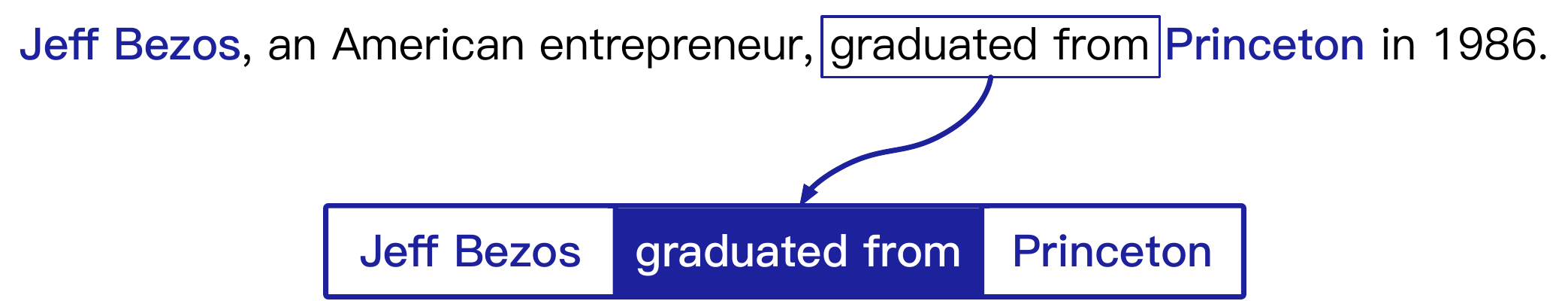}
    \caption{An example of open information extraction, which extracts relation arguments (entities) and phrases without relying on any pre-defined relation types.}
    \label{fig:openie}
\end{figure}

\begin{figure}
    \centering
    \includegraphics[width = \SIZE\linewidth]{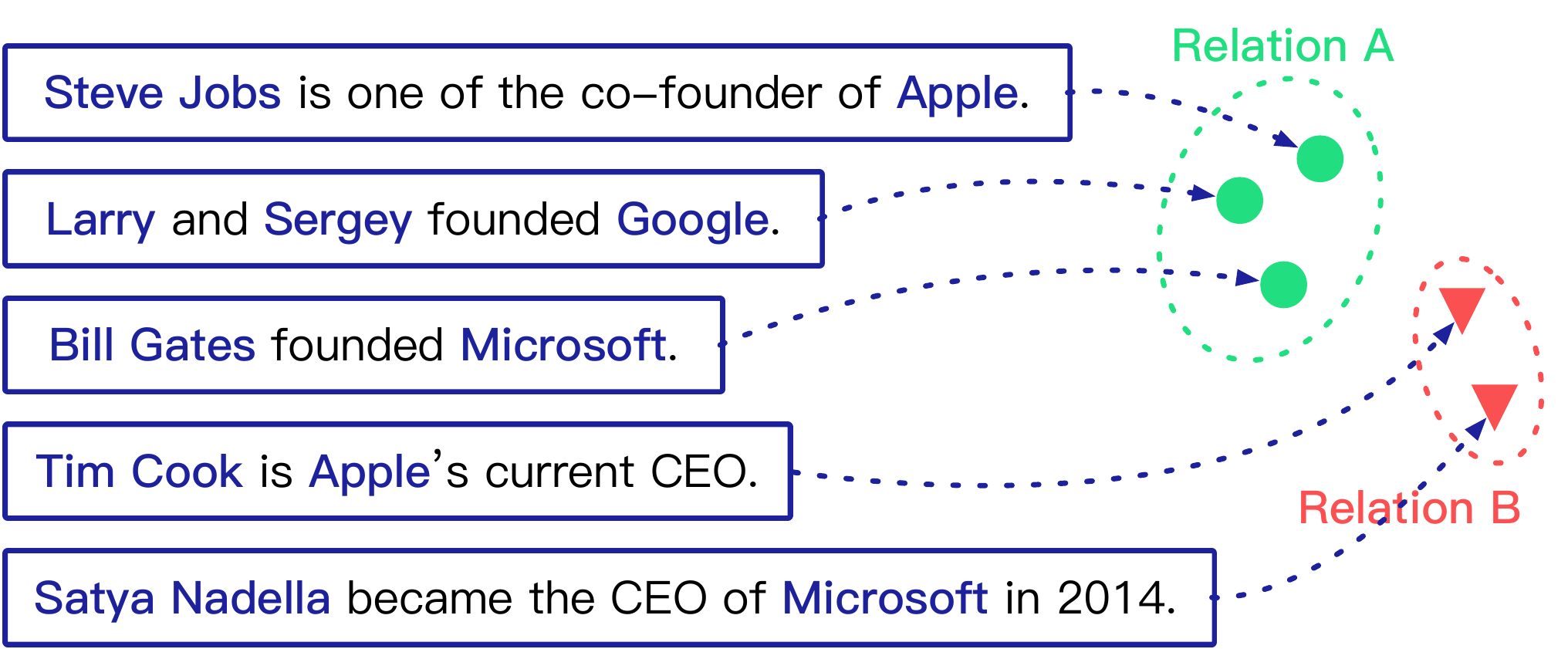}
    \caption{An example of clustering-based relation discovery, which identifying potential relation types by clustering unlabeled relational instances.}
    \label{fig:openre}
\end{figure}

\subsection{Orienting More Open Domains}
\label{sec:moreopen}

Most RE systems work within pre-specified relation sets designed by human experts. However, our world undergoes open-ended growth of relations and it is not possible to handle all these emerging relation types only by humans. Thus, we need RE systems that do not rely on pre-defined relation schemas and can work in open scenarios.

There are already some explorations in handling open relations: (1) \textbf{Open information extraction (Open~IE)}, as shown in Figure \ref{fig:openie}, extracts relation phrases and arguments (entities) from text \citep{banko2007open,fader-etal-2011-identifying,mausam-etal-2012-open,del2013clausie,angeli-etal-2015-leveraging,stanovsky-dagan-2016-creating,Mausam:2016:OIE:3061053.3061220,cui-etal-2018-neural}. Open~IE does not rely on specific relation types and thus can handle all kinds of relational facts. 
(2) \textbf{Relation discovery}, as shown in Figure \ref{fig:openre}, aims at discovering unseen relation types from unsupervised data. \citet{yao-etal-2011-structured,marcheggiani-titov-2016-discrete} propose to use generative models and treat these relations as latent variables, while \citet{shinyama2006preemptive, elsahar2017unsupervised, wu-etal-2019-open} cast relation discovery as a clustering task. 

Though relation extraction in open domains has been widely studied, there are still lots of unsolved research questions remained to be answered:

(1) \textbf{Canonicalizing relation phrases and arguments} in Open IE is crucial for downstream tasks \citep{niklaus-etal-2018-survey}. If not canonicalized, the extracted relational facts could be redundant and ambiguous. For example, Open~IE may extract two triples (\emph{Barack Obama}, \texttt{was born in}, \emph{Honolulu}) and (\emph{Obama}, \texttt{place of birth}, \emph{Honolulu}) indicating an identical fact. Thus, normalizing extracted results will largely benefit the applications of Open~IE. There are already some preliminary works in this area \citep{galarraga2014canonicalizing,Vashishth:2018:CCO:3178876.3186030} and more efforts are needed.

(2) The \textbf{not applicable (N/A) relation} has been hardly addressed in relation discovery. In previous work, it is usually assumed that the sentence always expresses a relation between the two entities \citep{marcheggiani-titov-2016-discrete}. However, in the real-world scenario, a large proportion of entity pairs appearing in a sentence do not have a relation, and ignoring them or using simple heuristics to get rid of them may lead to poor results. Thus, it would be of interest to study how to handle these N/A instances in relation discovery. 

\section{Other Challenges}

In this section, we analyze two key challenges faced by RE models, address them with experiments and show their significance in the research and development of RE systems\footnote{For more details about these experiments, please refer to our open-source toolkit~\url{https://github.com/thunlp/OpenNRE}.}.


\subsection{Learning from Text or Names}

\label{sec:name}

In the process of RE, both entity names and their context provide useful information for classification. 
\textbf{Entity names} provide typing information (e.g., we can easily tell \emph{JFK International Airport} is an airport) and help to narrow down the range of possible relations; In the training process, entity embeddings may also be formed to help relation classification (like in the link prediction task of KG). 
On the other hand, relations can usually be extracted from the semantics of \textbf{text} around entity pairs. In some cases, relations can only be inferred implicitly by reasoning over the context.

Since there are two sources of information, it is interesting to study how much each of them contributes to the RE performance. Therefore, we design three different settings for the experiments: (1) \textbf{normal} setting, where both names and text are taken as inputs; (2) \textbf{masked-entity} (ME) setting, where entity names are replaced with a special token; (3) \textbf{only-entity} (OE) setting, where only names of the two entities are provided. 



\begin{table}
    \small
    \centering
    \begin{tabular}{cccc}
        \toprule
        Benchmark & Normal & ME & OE \\
        \midrule
        Wiki80 (Acc) & 0.861 & 0.734 & 0.763 \\
        TACRED (F-1) & 0.666 & 0.554 & 0.412 \\
        NYT-10 (AUC) & 0.349 & 0.216 & 0.185 \\
        Wiki-Distant (AUC) & 0.222 & 0.145 & 0.173 \\
        \bottomrule
    \end{tabular}
    \caption{Results of state-of-the-arts models on the normal setting, masked-entity (ME) setting and only-entity (OE) setting. We report accuracies of BERT on Wiki80, F-1 scores of BERT on TACRED and AUC of PCNN-ATT on NYT-10 and Wiki-Distant. All models are from the OpenNRE package \citep{han-etal-2019-opennre}.}
    \label{tab:me}
\end{table}

Results from Table \ref{tab:me} show that compared to the normal setting, models suffer a huge performance drop in both the ME and OE settings. Besides, it is surprising to see that in some cases, only using entity names outperforms only using text with entities masked. It suggests that (1) both entity names and text provide crucial information for RE, and (2) for some existing state-of-the-art models and benchmarks, entity names contribute even more. 

The observation is contrary to human intuition: we classify the relations between given entities mainly from the text description, yet models learn more from their names. To make real progress in understanding how language expresses relational facts, this problem should be further investigated and more efforts are needed.  

\subsection{RE Datasets towards Special Interests}

\label{sec:special}

There are already many datasets that benefit RE research: For supervised RE, there are MUC \citep{grishman-sundheim-1996-message}, ACE-2005 \citep{NtroductionTheA2}, SemEval-2010 Task 8 \citep{hendrickx-etal-2009-semeval}, KBP37 \citep{zhang2015relation} and TACRED \citep{zhang2017position}; and we have NYT-10 \citep{riedel2010modeling}, FewRel \citep{han2018fewrel} and DocRED \citep{yao2019docred} for distant supervision, few-shot and document-level RE respectively. 

However, there are barely datasets targeting special problems of interest. For example, RE across sentences (e.g., two entities are mentioned in two different sentences) is an important problem, yet there is no specific datasets that can help researchers study it. 
Though existing document-level RE datasets contain instances of this case, it is hard to analyze the exact performance gain towards this specific aspect.
Usually, researchers (1) use hand-crafted sub-sets of general datasets or (2) carry out case studies to show the effectiveness of their models in specific problems, which is lacking of convincing and quantitative analysis. Therefore, to further study these problems of great importance in the development of RE, it is necessary for the community to construct well-recognized, well-designed and fine-grained datasets towards special interests. 


\section{Conclusion}

In this paper, we give a comprehensive and detailed review on the development of relation extraction models, generalize four promising directions leading to more powerful RE systems (utilizing more data, performing more efficient learning, handling more complicated context and orienting more open domains), and further investigate two key challenges faced by existing RE models. We thoroughly survey the previous RE literature as well as supporting our points with statistics and experiments. Through this paper, we hope to demonstrate the progress and problems in existing RE research and encourage more efforts in this area.

\section*{Acknowledgments}

This work is supported by the Natural Science Foundation of China (NSFC) and the German Research Foundation (DFG) in Project Crossmodal Learning, NSFC 61621136008 / DFG TRR-169, and Beijing Academy of Artificial Intelligence (BAAI). This work is also supported by the Pattern Recognition Center, WeChat AI, Tencent Inc. Gao is supported by 2019 Tencent Rhino-Bird Elite Training Program. Gao is also supported by Tsinghua University Initiative Scientific Research Program. 

\bibliography{aacl-ijcnlp2020}

\begin{thebibliography}{137}
\expandafter\ifx\csname natexlab\endcsname\relax\def\natexlab#1{#1}\fi

\bibitem[{Angeli et~al.(2015)Angeli, Johnson~Premkumar, and
  Manning}]{angeli-etal-2015-leveraging}
Gabor Angeli, Melvin~Jose Johnson~Premkumar, and Christopher~D. Manning. 2015.
\newblock \href {https://doi.org/10.3115/v1/P15-1034} {Leveraging linguistic
  structure for open domain information extraction}.
\newblock In \emph{Proceedings of ACL-IJCNLP}, pages 344--354.

\bibitem[{Bach and Badaskar(2007)}]{bach2007review}
Nguyen Bach and Sameer Badaskar. 2007.
\newblock \href
  {https://www.cs.cmu.edu/~nbach/papers/A-survey-on-Relation-Extraction.pdf} {A
  review of relation extraction}.

\bibitem[{Baldini~Soares et~al.(2019)Baldini~Soares, FitzGerald, Ling, and
  Kwiatkowski}]{soares2019matching}
Livio Baldini~Soares, Nicholas FitzGerald, Jeffrey Ling, and Tom Kwiatkowski.
  2019.
\newblock \href {https://doi.org/10.18653/v1/P19-1279} {Matching the blanks:
  Distributional similarity for relation learning}.
\newblock In \emph{Proceedings of ACL}, pages 2895--2905.

\bibitem[{Banko et~al.(2007)Banko, Cafarella, Soderland, Broadhead, and
  Etzioni}]{banko2007open}
Michele Banko, Michael~J Cafarella, Stephen Soderland, Matthew Broadhead, and
  Oren Etzioni. 2007.
\newblock \href {https://www.aaai.org/Papers/IJCAI/2007/IJCAI07-429.pdf} {Open
  information extraction from the web.}
\newblock In \emph{Proceedings of IJCAI}, pages 2670--2676.

\bibitem[{Beltagy et~al.(2019)Beltagy, Lo, and Ammar}]{beltagy2019combining}
Iz~Beltagy, Kyle Lo, and Waleed Ammar. 2019.
\newblock \href {https://www.aclweb.org/anthology/N19-1184.pdf} {Combining
  distant and direct supervision for neural relation extraction}.
\newblock In \emph{Proceedings of NAACL-HLT}, pages 1858--1867.

\bibitem[{Bordes et~al.(2014)Bordes, Chopra, and Weston}]{bordes2014question}
Antoine Bordes, Sumit Chopra, and Jason Weston. 2014.
\newblock \href {https://www.aclweb.org/anthology/D14-1067.pdf} {Question
  answering with subgraph embeddings}.
\newblock In \emph{Proceedings of EMNLP}, pages 615--620.

\bibitem[{Bordes et~al.(2013)Bordes, Usunier, Garcia-Duran, Weston, and
  Yakhnenko}]{bordes2013translating}
Antoine Bordes, Nicolas Usunier, Alberto Garcia-Duran, Jason Weston, and Oksana
  Yakhnenko. 2013.
\newblock \href
  {https://papers.nips.cc/paper/5071-translating-embeddings-for-modeling-multi-relational-data.pdf}
  {Translating embeddings for modeling multi-relational data}.
\newblock In \emph{Proceedings of NIPS}, pages 2787--2795.

\bibitem[{Bunescu and Mooney(2005)}]{bunescu2005shortest}
Razvan~C Bunescu and Raymond~J Mooney. 2005.
\newblock \href {https://www.aclweb.org/anthology/H05-1091.pdf} {A shortest
  path dependency kernel for relation extraction}.
\newblock In \emph{Proceedings of EMNLP}, pages 724--731.

\bibitem[{Cai et~al.(2016)Cai, Zhang, and Wang}]{cai2016bidirectional}
Rui Cai, Xiaodong Zhang, and Houfeng Wang. 2016.
\newblock \href {https://www.aclweb.org/anthology/P16-1072} {Bidirectional
  recurrent convolutional neural network for relation classification}.
\newblock In \emph{Proceedings of ACL}, pages 756--765.

\bibitem[{Califf and Mooney(1997)}]{califf-mooney-1997-relational}
Mary~Elaine Califf and Raymond~J. Mooney. 1997.
\newblock \href {https://www.aclweb.org/anthology/W97-1002} {Relational
  learning of pattern-match rules for information extraction}.
\newblock In \emph{Proceedings of CoNLL}, pages 9--15.

\bibitem[{Carlson et~al.(2010)Carlson, Betteridge, Kisiel, Settles, Hruschka,
  and Mitchell}]{carlson2010toward}
Andrew Carlson, Justin Betteridge, Bryan Kisiel, Burr Settles, Estevam~R
  Hruschka, and Tom~M Mitchell. 2010.
\newblock \href
  {https://www.aaai.org/ocs/index.php/AAAI/AAAI10/paper/view/1879/2201} {Toward
  an architecture for never-ending language learning}.
\newblock In \emph{Proceedings of AAAI}, pages 1306--1313.

\bibitem[{Christopoulou et~al.(2018)Christopoulou, Miwa, and
  Ananiadou}]{christopoulou2018walk}
Fenia Christopoulou, Makoto Miwa, and Sophia Ananiadou. 2018.
\newblock \href {https://www.aclweb.org/anthology/P18-2014.pdf} {A walk-based
  model on entity graphs for relation extraction}.
\newblock In \emph{Proceedings of ACL}, pages 81--88.

\bibitem[{Cui et~al.(2018)Cui, Wei, and Zhou}]{cui-etal-2018-neural}
Lei Cui, Furu Wei, and Ming Zhou. 2018.
\newblock \href {https://doi.org/10.18653/v1/P18-2065} {Neural open information
  extraction}.
\newblock In \emph{Proceedings of ACL}, pages 407--413.

\bibitem[{Culotta and Sorensen(2004)}]{culotta2004dependency}
Aron Culotta and Jeffrey Sorensen. 2004.
\newblock \href {https://www.aclweb.org/anthology/P04-1054.pdf} {Dependency
  tree kernels for relation extraction}.
\newblock In \emph{Proceedings of ACL}, page 423.

\bibitem[{Del~Corro and Gemulla(2013)}]{del2013clausie}
Luciano Del~Corro and Rainer Gemulla. 2013.
\newblock \href {https://dl.acm.org/citation.cfm?id=2488420} {Clausie:
  clause-based open information extraction}.
\newblock In \emph{Proceedings of WWW}, pages 355--366.

\bibitem[{Devlin et~al.(2019)Devlin, Chang, Lee, and
  Toutanova}]{devlin2019bert}
Jacob Devlin, Ming-Wei Chang, Kenton Lee, and Kristina Toutanova. 2019.
\newblock \href {https://www.aclweb.org/anthology/N19-1423.pdf} {Bert:
  Pre-training of deep bidirectional transformers for language understanding}.
\newblock In \emph{Proceedings of NAACL-HLT}, pages 4171--4186.

\bibitem[{Dong et~al.(2015)Dong, Wei, Zhou, and Xu}]{dong2015question}
Li~Dong, Furu Wei, Ming Zhou, and Ke~Xu. 2015.
\newblock \href {https://www.aclweb.org/anthology/P15-1026.pdf} {Question
  answering over freebase with multi-column convolutional neural networks}.
\newblock In \emph{Proceedings of ACL-IJCNLP}, pages 260--269.

\bibitem[{Du et~al.(2018)Du, Han, Way, and Wan}]{du2018multi}
Jinhua Du, Jingguang Han, Andy Way, and Dadong Wan. 2018.
\newblock \href {https://www.aclweb.org/anthology/D18-1245.pdf} {Multi-level
  structured self-attentions for distantly supervised relation extraction}.
\newblock In \emph{Proceedings of EMNLP}, pages 2216--2225.

\bibitem[{Elsahar et~al.(2017)Elsahar, Demidova, Gottschalk, Gravier, and
  Laforest}]{elsahar2017unsupervised}
Hady Elsahar, Elena Demidova, Simon Gottschalk, Christophe Gravier, and
  Frederique Laforest. 2017.
\newblock \href {https://link.springer.com/chapter/10.1007/978-3-319-70407-4_3}
  {Unsupervised open relation extraction}.
\newblock In \emph{Proceedings of ESWC}, pages 12--16.

\bibitem[{Fader et~al.(2011)Fader, Soderland, and
  Etzioni}]{fader-etal-2011-identifying}
Anthony Fader, Stephen Soderland, and Oren Etzioni. 2011.
\newblock \href {https://www.aclweb.org/anthology/D11-1142} {Identifying
  relations for open information extraction}.
\newblock In \emph{Proceedings of EMNLP}, pages 1535--1545.

\bibitem[{Feng et~al.(2018)Feng, Huang, Zhao, Yang, and
  Zhu}]{feng2018reinforcement}
Jun Feng, Minlie Huang, Li~Zhao, Yang Yang, and Xiaoyan Zhu. 2018.
\newblock \href
  {https://www.aaai.org/ocs/index.php/AAAI/AAAI18/paper/download/17151/16140}
  {Reinforcement learning for relation classification from noisy data}.
\newblock In \emph{Proceedings of AAAI}, pages 5779--5786.

\bibitem[{Finn et~al.(2017)Finn, Abbeel, and Levine}]{finn2017model}
Chelsea Finn, Pieter Abbeel, and Sergey Levine. 2017.
\newblock \href {http://proceedings.mlr.press/v70/finn17a/finn17a.pdf}
  {Model-agnostic meta-learning for fast adaptation of deep networks}.
\newblock In \emph{Proceedings of ICML}, pages 1126--1135.

\bibitem[{Gal{\'a}rraga et~al.(2014)Gal{\'a}rraga, Heitz, Murphy, and
  Suchanek}]{galarraga2014canonicalizing}
Luis Gal{\'a}rraga, Geremy Heitz, Kevin Murphy, and Fabian~M Suchanek. 2014.
\newblock \href {https://dl.acm.org/citation.cfm?id=2662073} {Canonicalizing
  open knowledge bases}.
\newblock In \emph{Proceedings of CIKM}, pages 1679--1688.

\bibitem[{Gao et~al.(2019)Gao, Han, Zhu, Liu, Li, Sun, and
  Zhou}]{gao-etal-2019-fewrel}
Tianyu Gao, Xu~Han, Hao Zhu, Zhiyuan Liu, Peng Li, Maosong Sun, and Jie Zhou.
  2019.
\newblock \href {https://doi.org/10.18653/v1/D19-1649} {{F}ew{R}el 2.0: Towards
  more challenging few-shot relation classification}.
\newblock In \emph{Proceedings of EMNLP-IJCNLP}, pages 6251--6256.

\bibitem[{Gerber and Chai(2010)}]{gerber2010ACL}
Matthew Gerber and Joyce Chai. 2010.
\newblock \href {https://www.aclweb.org/anthology/P10-1160.pdf} {Beyond
  {NomBank}: A study of implicit arguments for nominal predicates}.
\newblock In \emph{Proceedings of ACL}, pages 1583--1592.

\bibitem[{Gormley et~al.(2015)Gormley, Yu, and Dredze}]{gormley2015improved}
Matthew~R Gormley, Mo~Yu, and Mark Dredze. 2015.
\newblock \href {https://www.aclweb.org/anthology/D15-1205.pdf} {Improved
  relation extraction with feature-rich compositional embedding models}.
\newblock In \emph{Proceedings of EMNLP}, pages 1774--1784.

\bibitem[{Grishman and Sundheim(1996)}]{grishman-sundheim-1996-message}
Ralph Grishman and Beth Sundheim. 1996.
\newblock \href {https://www.aclweb.org/anthology/C96-1079} {Message
  understanding conference- 6: A brief history}.
\newblock In \emph{Proceedings of COLING}, pages 466--471.

\bibitem[{Han et~al.(2019)Han, Gao, Yao, Ye, Liu, and
  Sun}]{han-etal-2019-opennre}
Xu~Han, Tianyu Gao, Yuan Yao, Deming Ye, Zhiyuan Liu, and Maosong Sun. 2019.
\newblock \href {https://doi.org/10.18653/v1/D19-3029} {{O}pen{NRE}: An open
  and extensible toolkit for neural relation extraction}.
\newblock In \emph{Proceedings of EMNLP-IJCNLP}, pages 169--174.

\bibitem[{Han et~al.(2018{\natexlab{a}})Han, Liu, and Sun}]{han2018denoising}
Xu~Han, Zhiyuan Liu, and Maosong Sun. 2018{\natexlab{a}}.
\newblock \href {https://arxiv.org/pdf/1805.10959.pdf} {Denoising distant
  supervision for relation extraction via instance-level adversarial training}.
\newblock \emph{arXiv preprint arXiv:1805.10959}.

\bibitem[{Han et~al.(2018{\natexlab{b}})Han, Liu, and Sun}]{han2018neural}
Xu~Han, Zhiyuan Liu, and Maosong Sun. 2018{\natexlab{b}}.
\newblock \href
  {https://www.aaai.org/ocs/index.php/AAAI/AAAI18/paper/view/16691/16013}
  {Neural knowledge acquisition via mutual attention between knowledge graph
  and text}.
\newblock In \emph{Proceedings of AAAI}, pages 4832--4839.

\bibitem[{Han et~al.(2018{\natexlab{c}})Han, Yu, Liu, Sun, and
  Li}]{han2018hierarchical}
Xu~Han, Pengfei Yu, Zhiyuan Liu, Maosong Sun, and Peng Li. 2018{\natexlab{c}}.
\newblock \href {https://www.aclweb.org/anthology/D18-1247} {Hierarchical
  relation extraction with coarse-to-fine grained attention}.
\newblock In \emph{Proceedings of EMNLP}, pages 2236--2245.

\bibitem[{Han et~al.(2018{\natexlab{d}})Han, Zhu, Yu, Wang, Yao, Liu, and
  Sun}]{han2018fewrel}
Xu~Han, Hao Zhu, Pengfei Yu, Ziyun Wang, Yuan Yao, Zhiyuan Liu, and Maosong
  Sun. 2018{\natexlab{d}}.
\newblock \href {https://www.aclweb.org/anthology/D18-1514} {Fewrel: A
  large-scale supervised few-shot relation classification dataset with
  state-of-the-art evaluation}.
\newblock In \emph{Proceedings of EMNLP}, pages 4803--4809.

\bibitem[{Hendrickx et~al.(2009)Hendrickx, Kim, Kozareva, Nakov,
  {\'O}~S{\'e}aghdha, Pad{\'o}, Pennacchiotti, Romano, and
  Szpakowicz}]{hendrickx-etal-2009-semeval}
Iris Hendrickx, Su~Nam Kim, Zornitsa Kozareva, Preslav Nakov, Diarmuid
  {\'O}~S{\'e}aghdha, Sebastian Pad{\'o}, Marco Pennacchiotti, Lorenza Romano,
  and Stan Szpakowicz. 2009.
\newblock \href {https://www.aclweb.org/anthology/W09-2415} {{S}em{E}val-2010
  task 8: Multi-way classification of semantic relations between pairs of
  nominals}.
\newblock In \emph{Proceedings of {SEW}-2009}, pages 94--99.

\bibitem[{Hoffmann et~al.(2011)Hoffmann, Zhang, Ling, Zettlemoyer, and
  Weld}]{hoffmann2011knowledge}
Raphael Hoffmann, Congle Zhang, Xiao Ling, Luke Zettlemoyer, and Daniel~S Weld.
  2011.
\newblock \href {https://www.aclweb.org/anthology/P11-1055.pdf}
  {Knowledge-based weak supervision for information extraction of overlapping
  relations}.
\newblock In \emph{Proceedings of ACL}, pages 541--550.

\bibitem[{Hu et~al.(2019)Hu, Zhang, Shi, Nie, Guan, and Yang}]{hu2019improving}
Linmei Hu, Luhao Zhang, Chuan Shi, Liqiang Nie, Weili Guan, and Cheng Yang.
  2019.
\newblock \href {https://www.aclweb.org/anthology/D19-1395.pdf} {Improving
  distantly-supervised relation extraction with joint label embedding}.
\newblock In \emph{Proceedings of EMNLP-IJCNLP}, pages 3812--3820.

\bibitem[{Huang and Wang(2017)}]{huang2017deep}
Yi~Yao Huang and William~Yang Wang. 2017.
\newblock \href {https://www.aclweb.org/anthology/D17-1191.pdf} {Deep residual
  learning for weakly-supervised relation extraction}.
\newblock In \emph{Proceedings of EMNLP}, pages 1803--1807.

\bibitem[{Huffman(1995)}]{huffman1995learning}
Scott~B Huffman. 1995.
\newblock \href {https://doi.org/10.1007/3-540-60925-3_51} {Learning
  information extraction patterns from examples}.
\newblock In \emph{Proceedings of IJCAI}, pages 246--260.

\bibitem[{Ji et~al.(2017)Ji, Liu, He, Zhao et~al.}]{ji2017distant}
Guoliang Ji, Kang Liu, Shizhu He, Jun Zhao, et~al. 2017.
\newblock \href
  {https://www.aaai.org/ocs/index.php/AAAI/AAAI17/paper/viewPaper/14491}
  {Distant supervision for relation extraction with sentence-level attention
  and entity descriptions}.
\newblock In \emph{AAAI}, pages 3060--3066.

\bibitem[{Jiang and Zhai(2007)}]{jiang2007systematic}
Jing Jiang and ChengXiang Zhai. 2007.
\newblock \href {https://www.aclweb.org/anthology/N07-1015.pdf} {A systematic
  exploration of the feature space for relation extraction}.
\newblock In \emph{Proceedings of NAACL}, pages 113--120.

\bibitem[{Jiang et~al.(2017)Jiang, Shang, Cassidy, Ren, Kaplan, Hanratty, and
  Han}]{jiang2017metapad}
Meng Jiang, Jingbo Shang, Taylor Cassidy, Xiang Ren, Lance~M Kaplan, Timothy~P
  Hanratty, and Jiawei Han. 2017.
\newblock \href {https://dl.acm.org/citation.cfm?id=3098105} {Metapad: Meta
  pattern discovery from massive text corpora}.
\newblock In \emph{Proceedings of KDD}, pages 877--886.

\bibitem[{Kambhatla(2004)}]{kambhatla2004combining}
Nanda Kambhatla. 2004.
\newblock \href {https://www.aclweb.org/anthology/P04-3022.pdf} {Combining
  lexical, syntactic, and semantic features with maximum entropy models for
  extracting relations}.
\newblock In \emph{Proceedings of ACL}, pages 178--181.

\bibitem[{Kim and Moldovan(1995)}]{kim1995acquisition}
Jun-Tae Kim and Dan~I. Moldovan. 1995.
\newblock \href {https://ieeexplore.ieee.org/abstract/document/469825/}
  {Acquisition of linguistic patterns for knowledge-based information
  extraction}.
\newblock \emph{TKDE}, 7(5):713--724.

\bibitem[{Koch et~al.(2015)Koch, Zemel, and Salakhutdinov}]{koch2015siamese}
Gregory Koch, Richard Zemel, and Ruslan Salakhutdinov. 2015.
\newblock \href {http://www.cs.toronto.edu/~gkoch/files/msc-thesis.pdf}
  {Siamese neural networks for one-shot image recognition}.
\newblock In \emph{Proceedings of the Workshop of ICML}.

\bibitem[{Li et~al.(2016)Li, Sun, Johnson, Sciaky, Wei, Leaman, Davis,
  Mattingly, Wiegers, and Lu}]{li2016biocreative}
Jiao Li, Yueping Sun, Robin~J. Johnson, Daniela Sciaky, Chih{-}Hsuan Wei,
  Robert Leaman, Allan~Peter Davis, Carolyn~J. Mattingly, Thomas~C. Wiegers,
  and Zhiyong Lu. 2016.
\newblock \href {https://doi.org/10.1093/database/baw068} {{BioCreative} {V}
  {CDR} task corpus: a resource for chemical disease relation extraction}.
\newblock \emph{Database}, pages 1--10.

\bibitem[{Li et~al.(2020)Li, Long, Shen, Zhou, Yao, Huo, and
  Jiang}]{li2020self}
Yang Li, Guodong Long, Tao Shen, Tianyi Zhou, Lina Yao, Huan Huo, and Jing
  Jiang. 2020.
\newblock \href {https://www.aaai.org/Papers/AAAI/2020GB/AAAI-LiY.3203.pdf}
  {Self-attention enhanced selective gate with entity-aware embedding for
  distantly supervised relation extraction.}
\newblock In \emph{Proceedings of AAAI}, pages 8269--8276.

\bibitem[{Lin et~al.(2017)Lin, Liu, and Sun}]{lin2017neural}
Yankai Lin, Zhiyuan Liu, and Maosong Sun. 2017.
\newblock \href {https://www.aclweb.org/anthology/P17-1004.pdf} {Neural
  relation extraction with multi-lingual attention}.
\newblock In \emph{Proceedings of ACL}, pages 34--43.

\bibitem[{Lin et~al.(2015)Lin, Liu, Sun, Liu, and Zhu}]{lin2015learning}
Yankai Lin, Zhiyuan Liu, Maosong Sun, Yang Liu, and Xuan Zhu. 2015.
\newblock \href
  {https://www.aaai.org/ocs/index.php/AAAI/AAAI15/paper/download/9571/9523}
  {Learning entity and relation embeddings for knowledge graph completion}.
\newblock In \emph{Proceedings of AAAI}, pages 2181--2187.

\bibitem[{Lin et~al.(2016)Lin, Shen, Liu, Luan, and Sun}]{lin2016neural}
Yankai Lin, Shiqi Shen, Zhiyuan Liu, Huanbo Luan, and Maosong Sun. 2016.
\newblock \href {https://www.aclweb.org/anthology/P16-1200v2.pdf} {Neural
  relation extraction with selective attention over instances}.
\newblock In \emph{Proceedings of ACL}, pages 2124--2133.

\bibitem[{Liu et~al.(2013)Liu, Sun, Chao, and Che}]{liu2013convolution}
Chunyang Liu, Wenbo Sun, Wenhan Chao, and Wanxiang Che. 2013.
\newblock \href
  {https://link.springer.com/content/pdf/10.1007%2F978-3-642-53917-6.pdf}
  {Convolution neural network for relation extraction}.
\newblock In \emph{Proceedings of ICDM}, pages 231--242.

\bibitem[{Liu et~al.(2017)Liu, Wang, Chang, and Sui}]{liu2017soft}
Tianyu Liu, Kexiang Wang, Baobao Chang, and Zhifang Sui. 2017.
\newblock \href {https://www.aclweb.org/anthology/D17-1189} {A soft-label
  method for noise-tolerant distantly supervised relation extraction}.
\newblock In \emph{Proceedings of EMNLP}, pages 1790--1795.

\bibitem[{Liu et~al.(2015)Liu, Wei, Li, Ji, Zhou, and
  Houfeng}]{liu2015dependency}
Yang Liu, Furu Wei, Sujian Li, Heng Ji, Ming Zhou, and WANG Houfeng. 2015.
\newblock \href {https://www.aclweb.org/anthology/P15-2047.pdf} {A
  dependency-based neural network for relation classification}.
\newblock In \emph{Proceedings of ACL-IJCNLP}, pages 285--290.

\bibitem[{Marcheggiani and Titov(2016)}]{marcheggiani-titov-2016-discrete}
Diego Marcheggiani and Ivan Titov. 2016.
\newblock \href {https://doi.org/10.1162/tacl_a_00095} {Discrete-state
  variational autoencoders for joint discovery and factorization of relations}.
\newblock \emph{TACL}, 4:231--244.

\bibitem[{{Mausam} et~al.(2012){Mausam}, Schmitz, Soderland, Bart, and
  Etzioni}]{mausam-etal-2012-open}
{Mausam}, Michael Schmitz, Stephen Soderland, Robert Bart, and Oren Etzioni.
  2012.
\newblock \href {https://www.aclweb.org/anthology/D12-1048} {Open language
  learning for information extraction}.
\newblock In \emph{Proceedings of EMNLP-CoNLL}, pages 523--534.

\bibitem[{Mausam(2016)}]{Mausam:2016:OIE:3061053.3061220}
Mausam Mausam. 2016.
\newblock \href {http://dl.acm.org/citation.cfm?id=3061053.3061220} {Open
  information extraction systems and downstream applications}.
\newblock In \emph{Proceedings of IJCAI}, pages 4074--4077.

\bibitem[{Mikolov et~al.(2013{\natexlab{a}})Mikolov, Chen, Corrado, and
  Dean}]{mikolov2013efficient}
Tomas Mikolov, Kai Chen, Greg Corrado, and Jeffrey Dean. 2013{\natexlab{a}}.
\newblock \href {https://openreview.net/pdf?id=idpCdOWtqXd60} {Efficient
  estimation of word representations in vector space}.
\newblock In \emph{Proceedings of ICLR}.

\bibitem[{Mikolov et~al.(2013{\natexlab{b}})Mikolov, Sutskever, Chen, Corrado,
  and Dean}]{mikolov2013distributed}
Tomas Mikolov, Ilya Sutskever, Kai Chen, Greg~S Corrado, and Jeff Dean.
  2013{\natexlab{b}}.
\newblock \href
  {https://papers.nips.cc/paper/5021-distributed-representations-of-words-and-phrases-and-their-compositionality.pdf}
  {Distributed representations of words and phrases and their
  compositionality}.
\newblock In \emph{Proceedings of NIPS}, pages 3111--3119.

\bibitem[{Min et~al.(2013)Min, Grishman, Wan, Wang, and
  Gondek}]{min2013distant}
Bonan Min, Ralph Grishman, Li~Wan, Chang Wang, and David Gondek. 2013.
\newblock \href {https://www.aclweb.org/anthology/N13-1095.pdf} {Distant
  supervision for relation extraction with an incomplete knowledge base}.
\newblock In \emph{Proceedings of NAACL}, pages 777--782.

\bibitem[{Mintz et~al.(2009)Mintz, Bills, Snow, and
  Jurafsky}]{mintz2009distant}
Mike Mintz, Steven Bills, Rion Snow, and Dan Jurafsky. 2009.
\newblock \href {https://www.aclweb.org/anthology/P09-1113.pdf} {Distant
  supervision for relation extraction without labeled data}.
\newblock In \emph{Proceedings of ACL-IJCNLP}, pages 1003--1011.

\bibitem[{Mishra et~al.(2018)Mishra, Rohaninejad, Chen, and
  Abbeel}]{mishra2018simple}
Nikhil Mishra, Mostafa Rohaninejad, Xi~Chen, and Pieter Abbeel. 2018.
\newblock \href {https://openreview.net/forum?id=B1DmUzWAW} {A simple neural
  attentive meta-learner}.
\newblock In \emph{Proceedings of ICLR}.

\bibitem[{Miwa and Bansal(2016)}]{miwa2016end}
Makoto Miwa and Mohit Bansal. 2016.
\newblock \href {https://www.aclweb.org/anthology/P16-1105} {End-to-end
  relation extraction using lstms on sequences and tree structures}.
\newblock In \emph{Proceedings of ACL}, pages 1105--1116.

\bibitem[{Mooney and Bunescu(2006)}]{mooney2006subsequence}
Raymond~J Mooney and Razvan~C Bunescu. 2006.
\newblock \href
  {https://papers.nips.cc/paper/2787-subsequence-kernels-for-relation-extraction.pdf}
  {Subsequence kernels for relation extraction}.
\newblock In \emph{Proceedings of NIPS}, pages 171--178.

\bibitem[{Nakashole et~al.(2012)Nakashole, Weikum, and
  Suchanek}]{nakashole-etal-2012-patty}
Ndapandula Nakashole, Gerhard Weikum, and Fabian Suchanek. 2012.
\newblock \href {https://www.aclweb.org/anthology/D12-1104} {{PATTY}: A
  taxonomy of relational patterns with semantic types}.
\newblock In \emph{Proceedings of EMNLP-CoNLL}, pages 1135--1145.

\bibitem[{Narasimhan et~al.(2016)Narasimhan, Yala, and
  Barzilay}]{narasimhan2016improving}
Karthik Narasimhan, Adam Yala, and Regina Barzilay. 2016.
\newblock \href {https://www.aclweb.org/anthology/D16-1261.pdf} {Improving
  information extraction by acquiring external evidence with reinforcement
  learning}.
\newblock In \emph{Proceedings of EMNLP}, pages 2355--2365.

\bibitem[{Nguyen et~al.(2007)Nguyen, Matsuo, and Ishizuka}]{nguyen2007relation}
Dat~PT Nguyen, Yutaka Matsuo, and Mitsuru Ishizuka. 2007.
\newblock \href {https://www.aaai.org/Papers/AAAI/2007/AAAI07-224.pdf}
  {Relation extraction from wikipedia using subtree mining}.
\newblock In \emph{Proceedings of AAAI}, pages 1414--1420.

\bibitem[{Nguyen and Grishman(2015{\natexlab{a}})}]{nguyen2015combining}
Thien~Huu Nguyen and Ralph Grishman. 2015{\natexlab{a}}.
\newblock \href {https://arxiv.org/abs/1511.05926} {Combining neural networks
  and log-linear models to improve relation extraction}.
\newblock \emph{arXiv preprint arXiv:1511.05926}.

\bibitem[{Nguyen and Grishman(2015{\natexlab{b}})}]{nguyen2015relation}
Thien~Huu Nguyen and Ralph Grishman. 2015{\natexlab{b}}.
\newblock \href {https://www.aclweb.org/anthology/W15-1506} {Relation
  extraction: Perspective from convolutional neural networks}.
\newblock In \emph{Proceedings of the Workshop of NAACL}, pages 39--48.

\bibitem[{Nguyen and Moschitti(2011)}]{nguyen2011end}
Truc-Vien~T Nguyen and Alessandro Moschitti. 2011.
\newblock \href {https://www.aclweb.org/anthology/P11-2048.pdf} {End-to-end
  relation extraction using distant supervision from external semantic
  repositories}.
\newblock In \emph{Proceedings of ACL}, pages 277--282.

\bibitem[{Niklaus et~al.(2018)Niklaus, Cetto, Freitas, and
  Handschuh}]{niklaus-etal-2018-survey}
Christina Niklaus, Matthias Cetto, Andr{\'e} Freitas, and Siegfried Handschuh.
  2018.
\newblock \href {https://www.aclweb.org/anthology/C18-1326} {A survey on open
  information extraction}.
\newblock In \emph{Proceedings of COLING}, pages 3866--3878.

\bibitem[{Ntroduction(2005)}]{NtroductionTheA2}
Ii.~I Ntroduction. 2005.
\newblock \href
  {https://www.semanticscholar.org/paper/The-ACE-2005-(-ACE-05-)-Evaluation-Plan-Evaluation-Ntroduction/3a9b136ca1ab91592df36f148ef16095f74d009e}
  {The ace 2005 ( ace 05 ) evaluation plan evaluation of the detection and
  recognition of ace entities, values, temporal expression, relations, and
  events}.

\bibitem[{Pawar et~al.(2017)Pawar, Palshikar, and
  Bhattacharyya}]{pawar2017relation}
Sachin Pawar, Girish~K Palshikar, and Pushpak Bhattacharyya. 2017.
\newblock \href {https://arxiv.org/abs/1712.05191} {Relation extraction: A
  survey}.
\newblock \emph{arXiv preprint arXiv:1712.05191}.

\bibitem[{Peng et~al.(2017)Peng, Poon, Quirk, Toutanova, and
  Yih}]{peng2017nary}
Nanyun Peng, Hoifung Poon, Chris Quirk, Kristina Toutanova, and Wen-tau Yih.
  2017.
\newblock \href {https://transacl.org/ojs/index.php/tacl/article/view/1028}
  {Cross-sentence n-ary relation extraction with graph {LSTMs}}.
\newblock \emph{TACL}, 5:101--115.

\bibitem[{Qu et~al.(2019)Qu, Hua, Ouyang, Zhou, and Li}]{qu2019fine}
Jianfeng Qu, Wen Hua, Dantong Ouyang, Xiaofang Zhou, and Ximing Li. 2019.
\newblock \href {https://dl.acm.org/citation.cfm?id=3357384.3357997} {A
  fine-grained and noise-aware method for neural relation extraction}.
\newblock In \emph{Proceedings of CIKM}, pages 659--668.

\bibitem[{Quirk and Poon(2017)}]{quirk2017distant}
Chris Quirk and Hoifung Poon. 2017.
\newblock \href {http://www.aclweb.org/anthology/E17-1110} {Distant supervision
  for relation extraction beyond the sentence boundary}.
\newblock In \emph{Proceedings of EACL}, pages 1171--1182.

\bibitem[{Ravi and Larochelle(2017)}]{ravi2016optimization}
Sachin Ravi and Hugo Larochelle. 2017.
\newblock \href {https://openreview.net/pdf?id=rJY0-Kcll} {Optimization as a
  model for few-shot learning}.
\newblock In \emph{Proceedings of ICLR}.

\bibitem[{Riedel et~al.(2010)Riedel, Yao, and McCallum}]{riedel2010modeling}
Sebastian Riedel, Limin Yao, and Andrew McCallum. 2010.
\newblock \href {https://dl.acm.org/citation.cfm?id=1889799} {Modeling
  relations and their mentions without labeled text}.
\newblock In \emph{Proceedings of ECML-PKDD}, pages 148--163.

\bibitem[{Riedel et~al.(2013)Riedel, Yao, McCallum, and
  Marlin}]{riedel2013relation}
Sebastian Riedel, Limin Yao, Andrew McCallum, and Benjamin~M Marlin. 2013.
\newblock \href {https://www.aclweb.org/anthology/N13-1008.pdf} {Relation
  extraction with matrix factorization and universal schemas.}
\newblock In \emph{Proceedings of NAACL}, pages 74--84.

\bibitem[{Roth and Yih(2002)}]{roth2002probabilistic}
Dan Roth and Wen-tau Yih. 2002.
\newblock \href {https://www.aclweb.org/anthology/C02-1151} {Probabilistic
  reasoning for entity {\&} relation recognition}.
\newblock In \emph{Proceedings of COLING}.

\bibitem[{Roth and Yih(2004)}]{roth2004linear}
Dan Roth and Wen-tau Yih. 2004.
\newblock \href {https://www.aclweb.org/anthology/W04-2401.pdf} {A linear
  programming formulation for global inference in natural language tasks}.
\newblock In \emph{Proceedings of CoNLL}.

\bibitem[{Santos et~al.(2015)Santos, Xiang, and Zhou}]{santos2015classifying}
Cicero Nogueira~dos Santos, Bing Xiang, and Bowen Zhou. 2015.
\newblock \href {https://www.aclweb.org/anthology/P15-1061.pdf} {Classifying
  relations by ranking with convolutional neural networks}.
\newblock In \emph{Proceedings of ACL-IJCNLP}, pages 626--634.

\bibitem[{Sarawagi and Cohen(2005)}]{sarawagi2005semi}
Sunita Sarawagi and William~W Cohen. 2005.
\newblock \href
  {https://papers.nips.cc/paper/2648-semi-markov-conditional-random-fields-for-information-extraction}
  {Semi-markov conditional random fields for information extraction}.
\newblock In \emph{Proceedings of NIPS}, pages 1185--1192.

\bibitem[{Schlichtkrull et~al.(2018)Schlichtkrull, Kipf, Bloem, van~den Berg,
  Titov, and Welling}]{schlichtkrull2018modeling}
Michael Schlichtkrull, Thomas~N Kipf, Peter Bloem, Rianne van~den Berg, Ivan
  Titov, and Max Welling. 2018.
\newblock \href
  {https://link.springer.com/chapter/10.1007/978-3-319-93417-4_38} {Modeling
  relational data with graph convolutional networks}.
\newblock In \emph{Proceedings of ESWC}, pages 593--607.

\bibitem[{Shinyama and Sekine(2006)}]{shinyama2006preemptive}
Yusuke Shinyama and Satoshi Sekine. 2006.
\newblock \href {https://www.aclweb.org/anthology/N06-1039.pdf} {Preemptive
  information extraction using unrestricted relation discovery}.
\newblock In \emph{Proceedings of NAACL}, pages 304--311.

\bibitem[{Snell et~al.(2017)Snell, Swersky, and Zemel}]{snell2017prototypical}
Jake Snell, Kevin Swersky, and Richard Zemel. 2017.
\newblock \href
  {http://papers.nips.cc/paper/6996-prototypical-networks-for-few-shot-learning}
  {Prototypical networks for few-shot learning}.
\newblock In \emph{Proceedings of NIPS}, pages 4077--4087.

\bibitem[{Socher et~al.(2012)Socher, Huval, Manning, and
  Ng}]{socher2012semantic}
Richard Socher, Brody Huval, Christopher~D Manning, and Andrew~Y Ng. 2012.
\newblock \href {https://www.aclweb.org/anthology/D12-1110} {Semantic
  compositionality through recursive matrix-vector spaces}.
\newblock In \emph{Proceedings of EMNLP}, pages 1201--1211.

\bibitem[{Soderland et~al.(1995)Soderland, Fisher, Aseltine, and
  Lehnert}]{soderland1995crystal}
Stephen Soderland, David Fisher, Jonathan Aseltine, and Wendy Lehnert. 1995.
\newblock \href {https://www.ijcai.org/Proceedings/95-2/Papers/038.pdf}
  {Crystal inducing a conceptual dictionary}.
\newblock In \emph{Proceedings of IJCAI}, pages 1314--1319.

\bibitem[{Song et~al.(2018)Song, Zhang et~al.}]{song2018EMNLP}
Linfeng Song, Yue Zhang, et~al. 2018.
\newblock \href {https://www.aclweb.org/anthology/D18-1246} {N-ary relation
  extraction using graph-state lstm}.
\newblock In \emph{Proceedings of EMNLP}.

\bibitem[{Stanovsky and Dagan(2016)}]{stanovsky-dagan-2016-creating}
Gabriel Stanovsky and Ido Dagan. 2016.
\newblock \href {https://doi.org/10.18653/v1/D16-1252} {Creating a large
  benchmark for open information extraction}.
\newblock In \emph{Proceedings of EMNLP}, pages 2300--2305.

\bibitem[{Surdeanu et~al.(2012)Surdeanu, Tibshirani, Nallapati, and
  Manning}]{surdeanu2012multi}
Mihai Surdeanu, Julie Tibshirani, Ramesh Nallapati, and Christopher~D Manning.
  2012.
\newblock \href {https://www.aclweb.org/anthology/D12-1042.pdf} {Multi-instance
  multi-label learning for relation extraction}.
\newblock In \emph{Proceedings of EMNLP}, pages 455--465.

\bibitem[{Swampillai and Stevenson(2011)}]{swampillai2011RANLP}
Kumutha Swampillai and Mark Stevenson. 2011.
\newblock \href {https://www.aclweb.org/anthology/R11-1004} {Extracting
  relations within and across sentences}.
\newblock In \emph{Proceedings of RANLP}, pages 25–--32.

\bibitem[{Turian et~al.(2010)Turian, Ratinov, and Bengio}]{turian2010word}
Joseph Turian, Lev Ratinov, and Yoshua Bengio. 2010.
\newblock \href {https://aclweb.org/anthology/P10-1040} {Word representations:
  a simple and general method for semi-supervised learning}.
\newblock In \emph{Proceedings of ACL}, pages 384--394.

\bibitem[{Vashishth et~al.(2018)Vashishth, Jain, and
  Talukdar}]{Vashishth:2018:CCO:3178876.3186030}
Shikhar Vashishth, Prince Jain, and Partha Talukdar. 2018.
\newblock \href {https://doi.org/10.1145/3178876.3186030} {Cesi: Canonicalizing
  open knowledge bases using embeddings and side information}.
\newblock In \emph{Proceedings of WWW}, pages 1317--1327.

\bibitem[{Vaswani et~al.(2017)Vaswani, Shazeer, Parmar, Uszkoreit, Jones,
  Gomez, Kaiser, and Polosukhin}]{vaswani2017attention}
Ashish Vaswani, Noam Shazeer, Niki Parmar, Jakob Uszkoreit, Llion Jones,
  Aidan~N Gomez, {\L}ukasz Kaiser, and Illia Polosukhin. 2017.
\newblock \href
  {https://papers.nips.cc/paper/7181-attention-is-all-you-need.pdf} {Attention
  is all you need}.
\newblock In \emph{Proceedings of NIPS}, pages 5998--6008.

\bibitem[{Verga et~al.(2016)Verga, Belanger, Strubell, Roth, and
  McCallum}]{verga2016multilingual}
Patrick Verga, David Belanger, Emma Strubell, Benjamin Roth, and Andrew
  McCallum. 2016.
\newblock \href {https://www.aclweb.org/anthology/N16-1103} {Multilingual
  relation extraction using compositional universal schema}.
\newblock In \emph{Proceedings of NAACL}, pages 886--896.

\bibitem[{Verga and McCallum(2016)}]{vergamccallum}
Patrick Verga and Andrew McCallum. 2016.
\newblock \href {https://www.aclweb.org/anthology/W16-1312} {Row-less universal
  schema}.
\newblock In \emph{Proceedings of ACL}, pages 63--68.

\bibitem[{Verga et~al.(2018)Verga, Strubell, and
  McCallum}]{verga2018simultaneously}
Patrick Verga, Emma Strubell, and Andrew McCallum. 2018.
\newblock \href {https://www.aclweb.org/anthology/N18-1080} {Simultaneously
  self-attending to all mentions for full-abstract biological relation
  extraction}.
\newblock In \emph{Proceedings of NAACL-HLT}, pages 872--884.

\bibitem[{Vinyals et~al.(2016)Vinyals, Blundell, Lillicrap, Wierstra
  et~al.}]{vinyals2016matching}
Oriol Vinyals, Charles Blundell, Tim Lillicrap, Daan Wierstra, et~al. 2016.
\newblock \href
  {http://papers.nips.cc/paper/6385-matching-networks-for-one-shot-learning}
  {Matching networks for one shot learning}.
\newblock In \emph{Proceedings of NIPS}, pages 3630--3638.

\bibitem[{Vrande{\v{c}}i{\'c} and Kr{\"o}tzsch(2014)}]{vrandevcic2014wikidata}
Denny Vrande{\v{c}}i{\'c} and Markus Kr{\"o}tzsch. 2014.
\newblock \href {https://ai.google/research/pubs/pub42240} {Wikidata: a free
  collaborative knowledgebase}.
\newblock \emph{Proceedings of CACM}, 57(10):78--85.

\bibitem[{Vu et~al.(2016)Vu, Adel, Gupta et~al.}]{vu2016combining}
Ngoc~Thang Vu, Heike Adel, Pankaj Gupta, et~al. 2016.
\newblock \href {https://www.aclweb.org/anthology/N16-1065} {Combining
  recurrent and convolutional neural networks for relation classification}.
\newblock In \emph{Proceedings of NAACL}, pages 534--539.

\bibitem[{Wang et~al.(2016)Wang, Cao, De~Melo, and Liu}]{wang2016relation}
Linlin Wang, Zhu Cao, Gerard De~Melo, and Zhiyuan Liu. 2016.
\newblock \href {https://www.aclweb.org/anthology/P16-1123} {Relation
  classification via multi-level attention cnns}.
\newblock In \emph{Proceedings of ACL}, pages 1298--1307.

\bibitem[{Wang(2008)}]{wang2008re}
Mengqiu Wang. 2008.
\newblock \href {https://www.aclweb.org/anthology/I08-2119} {A re-examination
  of dependency path kernels for relation extraction}.
\newblock In \emph{Proceedings of IJCNLP}, pages 841--846.

\bibitem[{Wang et~al.(2018)Wang, Han, Lin, Liu, and Sun}]{wang2018adversarial}
Xiaozhi Wang, Xu~Han, Yankai Lin, Zhiyuan Liu, and Maosong Sun. 2018.
\newblock \href {https://www.aclweb.org/anthology/C18-1099} {Adversarial
  multi-lingual neural relation extraction}.
\newblock In \emph{Proceedings of COLING}, pages 1156--1166.

\bibitem[{Wang et~al.(2014)Wang, Zhang, Feng, and Chen}]{wang2014knowledge}
Zhen Wang, Jianwen Zhang, Jianlin Feng, and Zheng Chen. 2014.
\newblock \href
  {https://www.aaai.org/ocs/index.php/AAAI/AAAI14/paper/view/8531} {Knowledge
  graph embedding by translating on hyperplanes}.
\newblock In \emph{Proceedings of AAAI}, pages 1112--1119.

\bibitem[{Weston et~al.(2013)Weston, Bordes, Yakhnenko, and
  Usunier}]{weston2013connecting}
Jason Weston, Antoine Bordes, Oksana Yakhnenko, and Nicolas Usunier. 2013.
\newblock \href {https://www.aclweb.org/anthology/D13-1136} {Connecting
  language and knowledge bases with embedding models for relation extraction}.
\newblock In \emph{Proceedings of EMNLP}, pages 1366--1371.

\bibitem[{Wick et~al.(2006)Wick, Culotta et~al.}]{wick2006EMNLP}
Michael Wick, Aron Culotta, et~al. 2006.
\newblock \href {https://www.aclweb.org/anthology/W06-1671.pdf} {Learning field
  compatibilities to extract database records from unstructured text}.
\newblock In \emph{Proceedings of EMNLP}.

\bibitem[{Wu et~al.(2019)Wu, Yao, Han, Xie, Liu, Lin, Lin, and
  Sun}]{wu-etal-2019-open}
Ruidong Wu, Yuan Yao, Xu~Han, Ruobing Xie, Zhiyuan Liu, Fen Lin, Leyu Lin, and
  Maosong Sun. 2019.
\newblock \href {https://doi.org/10.18653/v1/D19-1021} {Open relation
  extraction: Relational knowledge transfer from supervised data to
  unsupervised data}.
\newblock In \emph{Proceedings of EMNLP-IJCNLP}, pages 219--228.

\bibitem[{Wu and He(2019)}]{wu2019enriching}
Shanchan Wu and Yifan He. 2019.
\newblock \href {https://dl.acm.org/doi/pdf/10.1145/3357384.3358119} {Enriching
  pre-trained language model with entity information for relation
  classification}.
\newblock In \emph{Proceedings of CIKM}, pages 2361--2364.

\bibitem[{Wu et~al.(2017)Wu, Bamman, and Russell}]{wu2017adversarial}
Yi~Wu, David Bamman, and Stuart Russell. 2017.
\newblock \href {https://www.aclweb.org/anthology/D17-1187} {Adversarial
  training for relation extraction}.
\newblock In \emph{Proceedings of EMNLP}, pages 1778--1783.

\bibitem[{Xiao and Liu(2016)}]{xiao2016semantic}
Minguang Xiao and Cong Liu. 2016.
\newblock \href {https://www.aclweb.org/anthology/C16-1119.pdf} {Semantic
  relation classification via hierarchical recurrent neural network with
  attention}.
\newblock In \emph{Proceedings of COLING}, pages 1254--1263.

\bibitem[{Xiong et~al.(2017)Xiong, Power, and Callan}]{xiong2017explicit}
Chenyan Xiong, Russell Power, and Jamie Callan. 2017.
\newblock \href {https://dl.acm.org/citation.cfm?id=3052558} {Explicit semantic
  ranking for academic search via knowledge graph embedding}.
\newblock In \emph{Proceedings of WWW}, pages 1271--1279.

\bibitem[{Xu et~al.(2015{\natexlab{a}})Xu, Feng, Huang, and
  Zhao}]{xu2015semantic}
Kun Xu, Yansong Feng, Songfang Huang, and Dongyan Zhao. 2015{\natexlab{a}}.
\newblock \href {http://www.emnlp2015.org/proceedings/EMNLP/pdf/EMNLP062.pdf}
  {Semantic relation classification via convolutional neural networks with
  simple negative sampling}.
\newblock In \emph{Proceedings of EMNLP}, pages 536--540.

\bibitem[{Xu et~al.(2016)Xu, Jia, Mou, Li, Chen, Lu, and Jin}]{xu2016improved}
Yan Xu, Ran Jia, Lili Mou, Ge~Li, Yunchuan Chen, Yangyang Lu, and Zhi Jin.
  2016.
\newblock \href {https://arxiv.org/abs/1601.03651} {Improved relation
  classification by deep recurrent neural networks with data augmentation}.
\newblock In \emph{Proceedings of COLING}, pages 1461--1470.

\bibitem[{Xu et~al.(2015{\natexlab{b}})Xu, Mou, Li, Chen, Peng, and
  Jin}]{xu2015classifying}
Yan Xu, Lili Mou, Ge~Li, Yunchuan Chen, Hao Peng, and Zhi Jin.
  2015{\natexlab{b}}.
\newblock \href {https://www.aclweb.org/anthology/D15-1206} {Classifying
  relations via long short term memory networks along shortest dependency
  paths}.
\newblock In \emph{Proceedings of EMNLP}, pages 1785--1794.

\bibitem[{Yao et~al.(2011)Yao, Haghighi, Riedel, and
  McCallum}]{yao-etal-2011-structured}
Limin Yao, Aria Haghighi, Sebastian Riedel, and Andrew McCallum. 2011.
\newblock \href {https://www.aclweb.org/anthology/D11-1135} {Structured
  relation discovery using generative models}.
\newblock In \emph{Proceedings of EMNLP}, pages 1456--1466.

\bibitem[{Yao et~al.(2019)Yao, Ye, Li, Han, Lin, Liu, Liu, Huang, Zhou, and
  Sun}]{yao2019docred}
Yuan Yao, Deming Ye, Peng Li, Xu~Han, Yankai Lin, Zhenghao Liu, Zhiyuan Liu,
  Lixin Huang, Jie Zhou, and Maosong Sun. 2019.
\newblock \href {https://doi.org/10.18653/v1/P19-1074} {{D}oc{RED}: A
  large-scale document-level relation extraction dataset}.
\newblock In \emph{Proceedings of ACL}, pages 764--777.

\bibitem[{Ye and Ling(2019)}]{ye-ling-2019-multi}
Zhi-Xiu Ye and Zhen-Hua Ling. 2019.
\newblock \href {https://doi.org/10.18653/v1/P19-1277} {Multi-level matching
  and aggregation network for few-shot relation classification}.
\newblock In \emph{Proceedings of ACL}, pages 2872--2881.

\bibitem[{Yoshikawa et~al.(2011)Yoshikawa, Riedel et~al.}]{yoshikawa2011}
Katsumasa Yoshikawa, Sebastian Riedel, et~al. 2011.
\newblock \href {https://link.springer.com/article/10.1186/2041-1480-2-S5-S6}
  {Coreference based event-argument relation extraction on biomedical text}.
\newblock \emph{Journal of Biomedical Semantics}, 2(S6).

\bibitem[{Yu and Lam(2010)}]{yu2010jointly}
Xiaofeng Yu and Wai Lam. 2010.
\newblock \href {https://www.aclweb.org/anthology/C10-2160} {Jointly
  identifying entities and extracting relations in encyclopedia text via a
  graphical model approach}.
\newblock In \emph{Proceedings of ACL}, pages 1399--1407.

\bibitem[{Zelenko et~al.(2003)Zelenko, Aone, and
  Richardella}]{zelenko2003kernel}
Dmitry Zelenko, Chinatsu Aone, and Anthony Richardella. 2003.
\newblock \href {http://www.jmlr.org/papers/volume3/zelenko03a/zelenko03a.pdf}
  {Kernel methods for relation extraction}.
\newblock \emph{Proceedings of JMLR}, pages 1083--1106.

\bibitem[{Zeng et~al.(2015)Zeng, Liu, Chen, and Zhao}]{zeng2015distant}
Daojian Zeng, Kang Liu, Yubo Chen, and Jun Zhao. 2015.
\newblock \href {http://www.emnlp2015.org/proceedings/EMNLP/pdf/EMNLP203.pdf}
  {Distant supervision for relation extraction via piecewise convolutional
  neural networks}.
\newblock In \emph{Proceedings of EMNLP}, pages 1753--1762.

\bibitem[{Zeng et~al.(2014)Zeng, Liu, Lai, Zhou, and Zhao}]{zeng2014relation}
Daojian Zeng, Kang Liu, Siwei Lai, Guangyou Zhou, and Jun Zhao. 2014.
\newblock \href {https://www.aclweb.org/anthology/C14-1220} {Relation
  classification via convolutional deep neural network}.
\newblock In \emph{Proceedings of COLING}, pages 2335--2344.

\bibitem[{Zeng et~al.(2017)Zeng, Lin, Liu, and Sun}]{zeng2016incorporating}
Wenyuan Zeng, Yankai Lin, Zhiyuan Liu, and Maosong Sun. 2017.
\newblock \href {https://www.aclweb.org/anthology/D17-1186} {Incorporating
  relation paths in neural relation extraction}.
\newblock In \emph{Proceedings of EMNLP}, pages 1768--1777.

\bibitem[{Zeng et~al.(2018)Zeng, He, Liu, and Zhao}]{zeng2018large}
Xiangrong Zeng, Shizhu He, Kang Liu, and Jun Zhao. 2018.
\newblock \href
  {https://www.aaai.org/ocs/index.php/AAAI/AAAI18/paper/viewPaper/16257} {Large
  scaled relation extraction with reinforcement learning}.
\newblock In \emph{Proceedings of AAAI}, pages 5658--5665.

\bibitem[{Zhang and Wang(2015)}]{zhang2015relation}
Dongxu Zhang and Dong Wang. 2015.
\newblock \href {https://arxiv.org/abs/1508.01006} {Relation classification via
  recurrent neural network}.
\newblock \emph{arXiv preprint arXiv:1508.01006}.

\bibitem[{Zhang et~al.(2006{\natexlab{a}})Zhang, Zhang, and
  Su}]{zhang2006exploring}
Min Zhang, Jie Zhang, and Jian Su. 2006{\natexlab{a}}.
\newblock \href {https://www.aclweb.org/anthology/N06-1037} {Exploring
  syntactic features for relation extraction using a convolution tree kernel}.
\newblock In \emph{Proceedings of NAACL}, pages 288--295.

\bibitem[{Zhang et~al.(2006{\natexlab{b}})Zhang, Zhang, Su, and
  Zhou}]{zhang2006composite}
Min Zhang, Jie Zhang, Jian Su, and Guodong Zhou. 2006{\natexlab{b}}.
\newblock \href {https://www.aclweb.org/anthology/P06-1104} {A composite kernel
  to extract relations between entities with both flat and structured
  features}.
\newblock In \emph{Proceedings of ACL}, pages 825--832.

\bibitem[{Zhang et~al.(2019{\natexlab{a}})Zhang, Deng, Sun, Wang, Chen, Zhang,
  and Chen}]{zhang2019long}
Ningyu Zhang, Shumin Deng, Zhanlin Sun, Guanying Wang, Xi~Chen, Wei Zhang, and
  Huajun Chen. 2019{\natexlab{a}}.
\newblock \href {https://www.aclweb.org/anthology/N19-1306} {Long-tail relation
  extraction via knowledge graph embeddings and graph convolution networks}.
\newblock In \emph{Proceedings of NAACL-HLT}, pages 3016--3025.

\bibitem[{Zhang et~al.(2015)Zhang, Zheng, Hu, and
  Yang}]{zhang2015bidirectional}
Shu Zhang, Dequan Zheng, Xinchen Hu, and Ming Yang. 2015.
\newblock \href {https://www.aclweb.org/anthology/Y15-1009} {Bidirectional long
  short-term memory networks for relation classification}.
\newblock In \emph{Proceedings of PACLIC}, pages 73--78.

\bibitem[{Zhang et~al.(2018)Zhang, Qi, and Manning}]{zhang2018graph}
Yuhao Zhang, Peng Qi, and Christopher~D. Manning. 2018.
\newblock \href {http://aclweb.org/anthology/D18-1244} {Graph convolution over
  pruned dependency trees improves relation extraction}.
\newblock In \emph{Proceedings of EMNLP}, pages 2205--2215.

\bibitem[{Zhang et~al.(2017)Zhang, Zhong, Chen, Angeli, and
  Manning}]{zhang2017position}
Yuhao Zhang, Victor Zhong, Danqi Chen, Gabor Angeli, and Christopher~D Manning.
  2017.
\newblock \href {https://nlp.stanford.edu/pubs/zhang2017tacred.pdf}
  {Position-aware attention and supervised data improve slot filling}.
\newblock In \emph{Proceedings of EMNLP}, pages 35--45.

\bibitem[{Zhang et~al.(2019{\natexlab{b}})Zhang, Han, Liu, Jiang, Sun, and
  Liu}]{zhang2019ernie}
Zhengyan Zhang, Xu~Han, Zhiyuan Liu, Xin Jiang, Maosong Sun, and Qun Liu.
  2019{\natexlab{b}}.
\newblock \href {https://doi.org/10.18653/v1/P19-1139} {{ERNIE}: Enhanced
  language representation with informative entities}.
\newblock In \emph{Proceedings of ACL}, pages 1441--1451.

\bibitem[{Zhao and Grishman(2005)}]{zhao2005extracting}
Shubin Zhao and Ralph Grishman. 2005.
\newblock \href {https://www.aclweb.org/anthology/P05-1052.pdf} {Extracting
  relations with integrated information using kernel methods}.
\newblock In \emph{Proceedings of ACL}, pages 419--426.

\bibitem[{Zheng et~al.(2019)Zheng, Han, Lin, Yu, Chen, Huang, Liu, and
  Xu}]{zheng-etal-2019-diag}
Shun Zheng, Xu~Han, Yankai Lin, Peilin Yu, Lu~Chen, Ling Huang, Zhiyuan Liu,
  and Wei Xu. 2019.
\newblock \href {https://doi.org/10.18653/v1/P19-1137} {{DIAG}-{NRE}: A neural
  pattern diagnosis framework for distantly supervised neural relation
  extraction}.
\newblock In \emph{Proceedings of ACL}, pages 1419--1429.

\bibitem[{Zhou et~al.(2005)Zhou, Su, Zhang, and Zhang}]{guodong2005exploring}
Guodong Zhou, Jian Su, Jie Zhang, and Min Zhang. 2005.
\newblock \href {https://www.aclweb.org/anthology/P05-1053} {Exploring various
  knowledge in relation extraction}.
\newblock In \emph{Proceedings of ACL}, pages 427--434.

\bibitem[{Zhou et~al.(2016)Zhou, Shi, Tian, Qi, Li, Hao, and
  Xu}]{zhou2016attention}
Peng Zhou, Wei Shi, Jun Tian, Zhenyu Qi, Bingchen Li, Hongwei Hao, and Bo~Xu.
  2016.
\newblock \href {https://www.aclweb.org/anthology/P16-2034} {Attention-based
  bidirectional long short-term memory networks for relation classification}.
\newblock In \emph{Proceedings of ACL}, pages 207--212.

\bibitem[{Zhu et~al.(2019{\natexlab{a}})Zhu, Lin, Liu, Fu, Chua, and
  Sun}]{zhu2019graph}
Hao Zhu, Yankai Lin, Zhiyuan Liu, Jie Fu, Tat-Seng Chua, and Maosong Sun.
  2019{\natexlab{a}}.
\newblock \href {https://doi.org/10.18653/v1/P19-1128} {Graph neural networks
  with generated parameters for relation extraction}.
\newblock In \emph{Proceedings of ACL}, pages 1331--1339.

\bibitem[{Zhu et~al.(2019{\natexlab{b}})Zhu, Deng, Xiong, Yu, Zhang, and
  Wang}]{zhu2019towards}
Mengdi Zhu, Zheye Deng, Wenhan Xiong, Mo~Yu, Ming Zhang, and William~Yang Wang.
  2019{\natexlab{b}}.
\newblock \href {https://arxiv.org/abs/1909.06058} {Towards open-domain named
  entity recognition via neural correction models}.
\newblock \emph{arXiv preprint arXiv:1909.06058}.

\bibitem[{Zhu et~al.(2019{\natexlab{c}})Zhu, Su, and Zhou}]{zhu2019improving}
Zhangdong Zhu, Jindian Su, and Yang Zhou. 2019{\natexlab{c}}.
\newblock \href {https://ieeexplore.ieee.org/document/8747447} {Improving
  distantly supervised relation classification with attention and semantic
  weight}.
\newblock \emph{IEEE Access}, 7:91160--91168.

\end{thebibliography}
\bibliographystyle{acl_natbib}

\end{document}